\documentclass[sigconf]{acmart}

\usepackage{algorithm}
\usepackage{subfigure}
\usepackage{multirow}
\usepackage[noend]{algpseudocode}
\AtBeginDocument{%
  \providecommand\BibTeX{{%
    \normalfont B\kern-0.5em{\scshape i\kern-0.25em b}\kern-0.8em\TeX}}}

\setcopyright{acmcopyright}
 
\begin{document}
\copyrightyear{2020}
\acmYear{2020}
\acmConference[WWW '20]{Proceedings of The Web Conference 2020}{April 20--24, 2020}{Taipei, Taiwan}
\acmBooktitle{Proceedings of The Web Conference 2020 (WWW '20), April 20--24, 2020, Taipei, Taiwan}
\acmPrice{}
\acmDOI{10.1145/3366423.3380089}
\acmISBN{978-1-4503-7023-3/20/04}
\title{Relation Adversarial Network for Low Resource Knowledge Graph Completion}
%\title{Relation-Gated Domain Adaptation  for   Relation Extraction}

\author{Ningyu Zhang}
\authornote{Equal contribution and shared co-first authorship.}
\email{ningyu.zny@alibaba-inc.com}
\affiliation{%
  \institution{Alibaba Group \& AZFT Joint Lab for Knowledge Engine}
  \country{China}
}
\author{Shumin Deng}
\authornotemark[1]
\email{231sm@zju.edu.cn}
\affiliation{%
  \institution{Zhejiang University \& AZFT Joint Lab for Knowledge Engine}
  \country{China}
}

\author{Zhanlin Sun}
\email{zhanlins@andrew.cmu.edu}
\affiliation{%
  \institution{School of Computer Science\\ Carnegie Mellon University}
  \country{United States}
}

\author{Jiaoyan Chen}
\email{jiaoyan.chen@cs.ox.ac.uk}
\affiliation{%
  \institution{Department of Computer Science\\ Oxford University}
  \country{United Kingdom}
}

\author{Wei Zhang}
\email{lantu.zw@alibaba-inc.com}
\affiliation{%
  \institution{Alibaba Group \& AZFT Joint Lab for Knowledge Engine}
  \country{China}
}

\author{Huajun Chen}
\authornote{Corresponding author.}
\email{huajunsir@zju.edu.cn}
\affiliation{%
  \institution{Zhejiang University \& AZFT Joint Lab for Knowledge Engine}
  \country{China}
}

% \renewcommand{\shortauthors}{Ningyu Zhang et al.}


\begin{abstract}
Knowledge Graph Completion (KGC) has been proposed to improve Knowledge Graphs by filling in missing connections via link prediction or relation extraction. One of the main difficulties for KGC is a low resource problem. Previous approaches assume sufficient training triples to learn versatile vectors for entities and relations, or a satisfactory number of labeled sentences to train a competent relation extraction model. However, low resource relations are very common in KGs, and those newly added relations often do not have many known samples for training. In this work, we aim at predicting new facts under a challenging setting where only limited training instances are available. We propose a general framework called \emph{Weighted Relation Adversarial Network}, which utilizes an adversarial procedure to help adapt knowledge/features learned from high resource relations to different but related low resource relations. Specifically, the framework takes advantage of a relation discriminator to distinguish between samples from different relations, and help learn relation-invariant features more transferable from source relations to target relations. Experimental results show that the proposed approach outperforms previous methods regarding low resource settings for both link prediction and relation extraction.
\end{abstract}


\maketitle

\section{Introduction}
Knowledge Graphs (KGs) organize facts in a structured way as triples in the form of \emph{<subject, predicate, object>}, abridged as $(s, p, o)$, where $s$ and $o$ denote entities and $p$ builds relations between entities. Due to the convenience of building semantic connections between data from various sources, many large-scale KGs have been built in recent years and have led to a broad range of important applications, such as question answering \cite{yu-etal-2017-improved}, language understanding\cite{zhang-etal-2019-ernie}, relational data analytics \cite{zhang-etal-2019-long}, recommender systems \cite{Rec:conf/www/WangZZLXG19}, etc.. However, these KGs are usually far from complete and noisy; thus various Knowledge Graph Completion (KGC) methods have been proposed in recent years. KGC could be achieved either by \emph{\textbf{Link Prediction}} via typical KG embedding models such as TransE \cite{TransE:conf/nips/BordesUGWY13}, DistMult \cite{DistMult:conf/iclr/2015}, etc., or by \emph{\textbf{Relation Extraction}} (RE) models which aim to complete missing relations by extracting new relational facts from textual corpus \cite{plank2013embedding,nguyen2014employing}.

\begin{figure}[!htbp]
\centering
\includegraphics [scale=0.19]{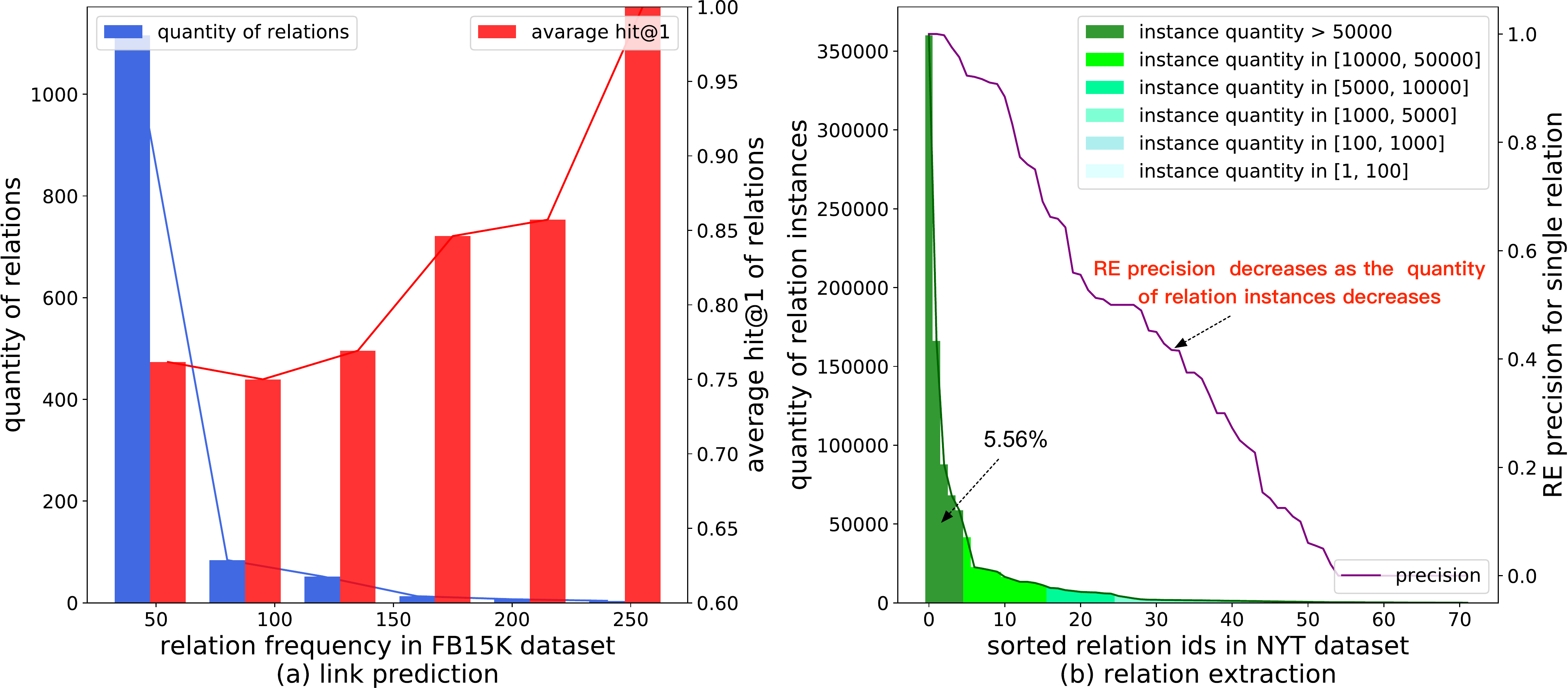}
\caption{Illustration of low resource problems in typical KGC tasks: (a) link prediction, and (b) relation extraction. The relation frequency in (a) refers to the number of triples existing in FB15K for a relation. The relations IDs are sorted by the number of labeled sentences for that relation in (b).} 
\label{fig:low_resource}
\end{figure}

\textbf{Long Tail and Low Resource Problems.}
One of the main difficulties for KGC is the long tail and low resource problems. For example, most KG embedding models for \emph{\textbf{Link Prediction}} assume sufficient training triples which are indispensable when learning versatile vectors for entities and relations. However, as illustrated in Figure~\ref{fig:low_resource}(a),  the blue line shows there is a large portion of relations in FB15K having only a few triples, revealing the common existence of \emph{long tail low resource relations}. The red line shows that the prediction results of relations are highly related to their frequency, and the results of low resource relations are much worse than those of highly frequent ones. \emph{\textbf{Relation Extraction}} suffers identical difficulty. For example, distant supervision (DS) methods were proposed to mitigate labeling laboriousness by aligning existing entity pairs of relation with textual sentences to generate training instances for that relation \cite{qin-etal-2018-dsgan} automatically. However, this requires both sufficient entity pairs of that relation existing in the KG and an adequate number of matching sentences in the textual corpus correspondingly, which are difficult to be achieved for most relations in many KGs.  As depicted in Figure~\ref{fig:low_resource}(b), similarly, the green line shows there is a large portion of relations in NYT-Freebase having relatively low numbers of labeled sentences, and the prediction results decrease severely for those long-tail relations.
%\dsm{only 5.56\% relations in the NYT-Freebase dataset have over extremely more pair-sentence alignment than others, and the prediction results are relatively low for those long-tail relations.}

% ===== in order to put the figure on the page where it first mentioned =====
\begin{figure*}[!htbp]
\centering
\includegraphics [scale=0.315]{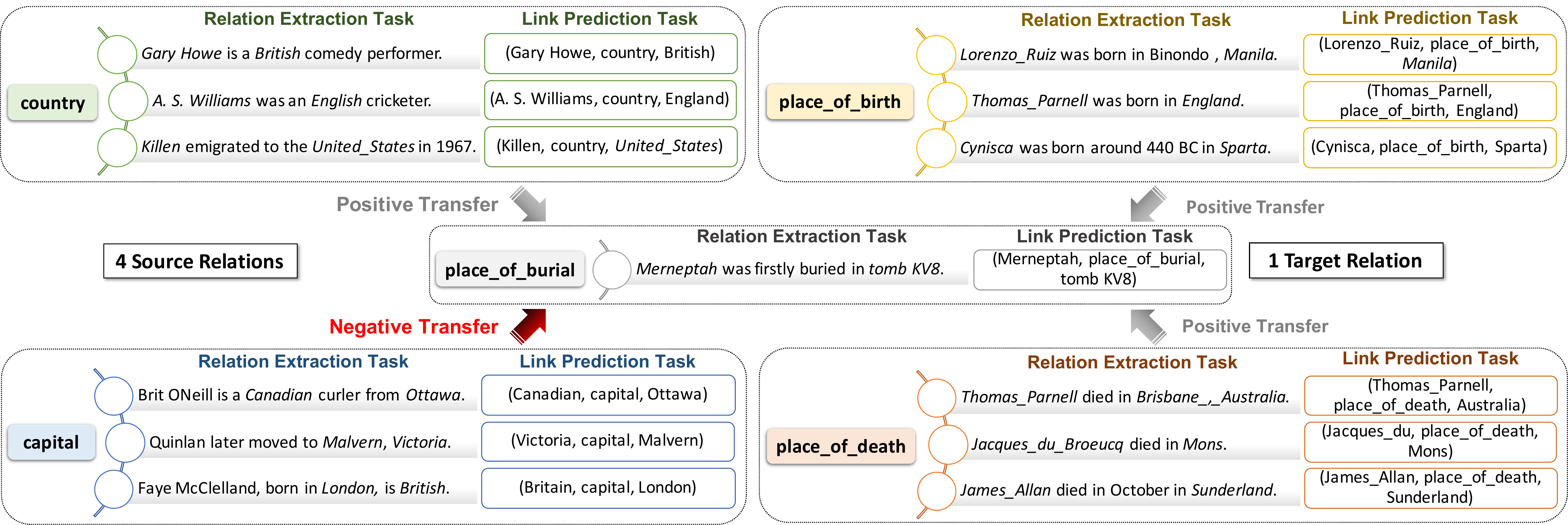}
\vspace{-3mm}
\caption{Illustration of relation adaptation where four source relations are adapted to one target relation. Note that there are two KGC tasks, relation extraction and link prediction, for each relation. In RE tasks, entity pairs are marked in italic.} 
% \vspace{-3mm}
\label{fig:ra}
\end{figure*}

\textbf{Adversarial Transfer Learning.}
Transfer learning is a research paradigm that focuses on transferring knowledge gained while learning from one domain, typically of high resource, and applying it to different but related domains, typically of low resource \cite{rosenstein2005transfer}. For example, the knowledge gained while learning to predict the relation \emph{place\_of\_death} could apply when trying to predict \emph{place\_of\_burial} if the latter lacks training samples. \emph{Domain Adversarial Network} (DAN) has been successfully applied in transfer learning and is among top-performance architectures in standard transfer and domain adaptation settings \cite{shu2019transferable}. The core idea is to extract domain-invariant features via an adversarial learning procedure that is capable of reducing the distribution discrepancy between the source domain and the target domain. Typically, domain adversarial networks consist of two adversarial players, where the first one is a domain discriminator $D$ trained to distinguish the source domain from the target domain, and the second one is a feature extractor $F$ which seeks to learn domain-invariant representations and serves as the adversary to confuse the domain discriminator \cite{shu2019transferable}. The final training convergence produces a feature extractor that excels in extracting more transferable features across domains.

\textbf{Relation Adversarial Network.}
Inspired by the idea of \emph{Domain Adversarial Networks} for domain adaptation, we propose \textbf{R}elation \textbf{A}dversarial \textbf{N}etwork (\textbf{RAN}) for relation adaptation, and raise the research question: \textbf{whether it is feasible to adapt knowledge/features learnt from high resource relations such as \emph{place\_of\_death},  to a different but related low resource relation, such as \emph{place\_of\_burial}, via an adversarial procedure?} Similarly, the \emph{relation adversarial network} consists of two adversarial players, where the first player is a relation discriminator $D$ trained to distinguish the source relation (i.e., the \emph{place\_of\_death}) from the target relation (i.e., the \emph{place\_of\_burial}). The second player is a feature extractor or instance encoder $F$, which tries to confuse the relation discriminator and seeks to learn common-features adaptable from source to target relations. The key intuition is to learn general relation-invariant features that can disentangle the factors of linguistic variations underlying different relations and close the linguistic gap between related relations. 

\textbf{Negative Transfer.}
Unlike standard adversarial domain adaptation, which assumes fully identical and shared label space between the source and target domains, it is expected for RAN to consider the adaptation from multiple source relations to one or multiple target relations. For example, in Figure~\ref{fig:ra}, the RAN model is expected to learn general location information from three source relations (\emph{place\_of\_death}, \emph{place\_of\_birth} and \emph{country}), and then adapt implied knowledge to the target relation (\emph{place\_of\_burial}) to improve its prediction performance. This new problem set is more practical and challenging than standard domain adaptation, since different relations might have different impacts on the transfer and there exist outlier source relations that might cause \emph{negative transfer} \cite{rosenstein2005transfer}  when discriminating from the target relation. As seen in Figure~\ref{fig:ra}, the source relation \emph{capital} may result in negative transfer to the target relation \emph{place\_of\_burial} as \emph{capital} does not describe a relation between a person and a place. Although \emph{partial transfer learning} has already been proposed to relax the shared label space assumption \cite{long2018, zhang2018importance}, it still requires that the target label space is a subspace of the source label space. In contrast, source relations could be completely different from target relation(s) in our problem settings, where it is even more critical to address the \emph{negative transfer} problem.

To address the aforementioned issues, we propose a general framework called \textbf{W}eighted \textbf{R}elation \textbf{A}dversarial \textbf{N}etwork (\textbf{wRAN}), which consists of three modules: 
(1) Instance encoder, which learns transferable features that can disentangle explanatory factors of \textbf{linguistic variations} cross relations. We implement the instance encoder with a convolutional neural network (CNN) considering both model performance and time efficiency. Other neural architectures such as recurrent neural networks (RNN) and BERT \cite{devlin2018bert} can also be used as encoders. 
(2) Adversarial relation adaptation, which looks for a relation discriminator that can distinguish among samples having different relation distributions. Adversarial learning helps learn a neural network that can map a target sample to a feature space such that the discriminator will no longer distinguish it from a source sample. 
(3) Weighed relation adaptation, which identifies unrelated source relations/samples and down-weight their importance automatically to tackle the problem of \emph{negative transfer} and encourage \emph{positive transfer}.

We evaluate wRAN's feasibility with two different KGC tasks: link prediction and relation extraction, both in low resource settings. They share the same adversarial learning framework but with varying types of instance encoders. The former encodes triples of a relation, and the latter learn a sentence feature extractor that can learn general relation-invariant textual features to close the linguistic gap between relations. 

The main contributions of the paper are summarized as follows.

\begin{itemize}
    \item To the best of our knowledge, we are the first to apply adversarial transfer learning in tackling the low resource problems for knowledge graph completion, which could be easily integrated with both link prediction and relation extraction.
    \item We propose a general framework called \textbf{R}elation \textbf{A}dversarial \textbf{N}etwork (RAN), which utilizes a relation discriminator to distinguish between samples from different relations and help learn relation-invariant features transferable from source relations to target relations. 
    \item We present a relation-gated mechanism that fully relaxes the shared label space assumptions. This mechanism selects out outlier source relations/samples and alleviates negative transfer of irrelevant relations/samples, which can be trained in an end-to-end framework. 
    \item Experiments show that our framework exceeds state-of-the-art results in low resource settings on public benchmark datasets for both link prediction and relation extraction.
\end{itemize}

\section{Related Work}
\textbf{Knowledge Graph Completion.} 
KGC can be achieved either by link prediction from knowledge graph \cite{IterE:conf/www/ZhangPWCZZBC19} or  by extracting new relational facts from textual corpus \cite{zhang-etal-2019-long}. A variety of link prediction approaches have been proposed to encode entities and relations into a continuous low-dimensional space. TransE \cite{TransE:conf/nips/BordesUGWY13}  regards the relation $r$ in the given fact $(h, r, t)$ as a translation from $h$ to $t$ within the low-dimensional space. TransE achieves good results and has many extensions, including TransR \cite{TransR:conf/aaai/LinLSLZ15}, TransD \cite{TransD:conf/acl/JiHXL015}, TransH \cite{TransH:conf/aaai/WangZFC14}, etc.  RESCAL \cite{RESCAL:conf/icml/NickelTK11}  studies on matrix factorization based knowledge graph embedding models using a bilinear form as score function.  DistMult \cite{DistMult:conf/iclr/2015} simplifies RESCAL by using a diagonal matrix to encode relation, and ComplEx \cite{ComplEx:conf/icml/TrouillonWRGB16} extends DistMult into the complex number field. ConvE \cite{ConvE:conf/aaai/DettmersMS018} uses a convolutions neural network as the score function.  Analogy \cite{ANALOGY:conf/icml/LiuWY17} optimize the latent representations with respect to the analogical properties of the embedded entities and relations.  KBGAN \cite{cai2017kbgan} generates  better negative examples to train knowledge graph embedding models via  an adversarial learning framework.   RotatE \cite{RotatE:Sun2019RotatE} defines each relation as a rotation from the source entity to the target entity in the complex vector space. KG-BERT \cite{KG-BERT:Yao2019KG} takes entity and relation descriptions of a triple as input and computes scoring function of the triple with BERT. For the low resource setting,  \cite{xiong2018one} proposed a one-shot relational learning framework, which learns a matching metric by considering both the learned embeddings and one-hop graph structures.  \cite{chen2019meta} proposed a Meta Relational Learning (MetaR) framework to do few-shot link prediction in KGs. \cite{IterE:conf/www/ZhangPWCZZBC19} propose a novel framework IterE iteratively learning embeddings and rules which can improve the quality of sparse entity embeddings and their link prediction results. 

Relation extraction aims to detect and categorize semantic relations between a pair of entities. To alleviate the annotations given by human experts, weak supervision and distant supervision have been employed to automatically generate annotations based on KGs (or seed patterns/instances) \cite{zeng2015distant,lin2016neural,ji2017distant,he2018see,zhang2018capsule,zeng2018large,qin-etal-2018-dsgan,zhang-etal-2019-long,zhang2019transfer}.    However, all these models merely focus on extracting facts from a single domain, ignoring the rich information in other domains. Recently, there have been only a few studies on low resource relation extraction \cite{plank2013embedding,nguyen2014robust,nguyen2014employing,nguyen2015semantic,fu2017domain}. Of these, \cite{nguyen2014robust} followed the supervised domain adaptation paradigm. In contrast, \cite{plank2013embedding,nguyen2014employing} worked on unsupervised domain adaptation. \cite{zhang2019long} proposes to take advantage of the knowledge from data-rich classes at the head of the distribution to boost the performance of the data-poor classes at the tail. \cite{fu2017domain,nguyen2015semantic} presented adversarial learning algorithms for unsupervised domain adaptation tasks. However, their methods suffer from the negative transfer bottleneck when encountered partial domain adaptation. Our approach relaxes the label constraints that source relations could be completely different from target relations, which is more general and useful in real scenarios. 
% rios2018generalizing

\textbf{Adversarial Domain Adaptation.}
Generative adversarial nets (GANs) \cite{goodfellow2014generative}  have become a popular solution to reduce domain discrepancy through an adversarial objective concerning a domain classifier \cite{ganin2016domain,tzeng2017adversarial,shen2018wasserstein}. Recently, only a few domain adaptation algorithms \cite{cao2017partial, chen2018re} that can handle  \textbf{imbalanced relation distribution} or  \textbf{partial adaptation} have been proposed. \cite{long2018} proposed a method to simultaneously alleviate negative transfer by down-weighting the data of outlier source classes in category level. \cite{zhang2018importance} proposed an adversarial nets-based partial domain adaptation method to identify the source samples at the instance level. However,  most of these studies concentrate on image classification. Different from images, the text is more diverse and nosier.  We believe these methods may transfer to the low resource knowledge graph completion, but the effect of exact modifications is not apparent. We make the very first attempt to investigate the empirical results of these methods.  Moreover, we propose a relation-gate mechanism to explicitly model both coarse-grained and fine-grained knowledge transfer to lower the negative transfer effects from categories and samples.

% ===== in order to put the figure on the page where it first mentioned =====
\begin{figure*}[!htbp]
\centering
\includegraphics [width=1\textwidth]{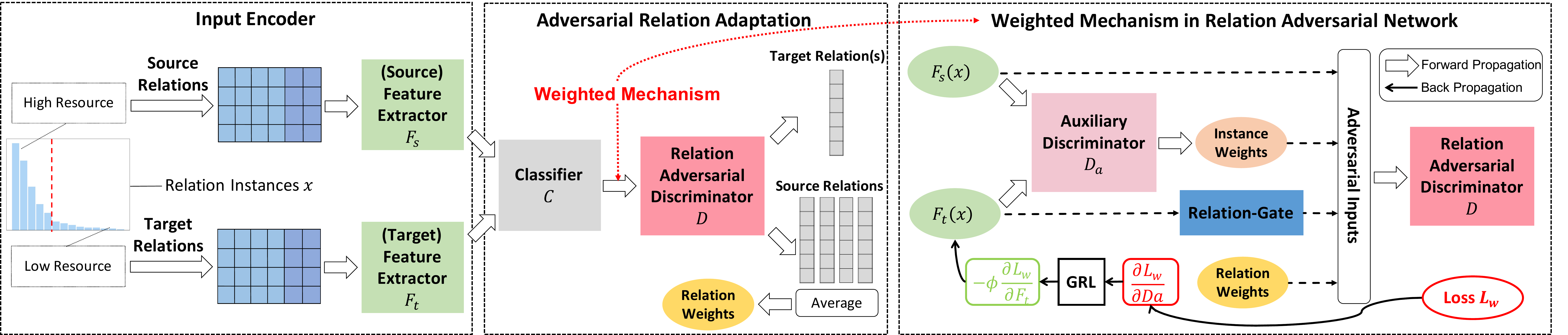}% 1\linewidth
\caption{Overview of our approach. 
%The parameters of the instance encoder for the source ($F_s$) and relation classifier $C$ are pre-learned and subsequently fixed. The yellow part denotes the probabilities of assigning the target relation to the source classifier to obtain \textbf{relation weights}. The orange part denotes the output probabilities of auxiliary relation discriminator to obtain \textbf{instance weights}.  The blue part denotes the \textbf{relation-gate}.  The red part denotes the relation adversarial classifier.  GRL \cite{ganin2016domain} denotes the gradient reversal layer.
} 
\label{fig:architecture}
\end{figure*}

\section{Methodology}
%In this section, we introduce the overall framework of our approach for long-tail knowledge graph completion, as shown in Figure~\ref{fig:architecture}. We begin by defining the problem.

\subsection{Problem Definition}

In this paper, we propose \emph{relation adaptation} where we allow adaptation from multiple source relations to one or multiple target relations. This flexible paradigm finds real applications in practice, as we could perform transferring from a number of high resource relations to a single or a few long-tail low resource relations.

Being similar but slightly different to standard domain adaptation, in relation adaptation, we are provided with a set of  \emph{source relations} $S=\{R_s^i\}_{i=1}^{r_s}$ with $r_s$ source relations, where for each relation we have $R_s^i=\{x_j^s,y_j^s\}_{j=1}^{n_s}$ with $n_s$ labeled examples. Given one target relation $R_t=\{x_j^t\}_{j=1}^{n_t}$ with $n_t$ unlabelled  samples, the goal is to predict the labels of samples in the low resource relation $R_t$ by adapting knowledge/features learnt from those high resource relations $\{R_s^i\}_{i=1}^{r_s}$.

Generally, the instances of source relation and target relation are sampled from probability distributions $p$ and $q$ respectively. Typically, we have $p \not=q$. The goal of this paper is to design a deep neural network   $\textbf{f}= F(x)$ that enables learning of transferable features from all other source relations and an adaptive classifier $y = G(\textbf{f})$ to bridge the cross-relation discrepancy, such that the risk for target relation prediction $Pr_{(x,y) \sim q}[G(F(x)) \not=y]$ is minimized by leveraging the source relation supervisions.

\subsection{Framework Overview}
 Our model consists of three parts, as shown in Figure~\ref{fig:architecture}:

\textbf{Instance Encoder.}
Given an instance of entity pairs or sentences, we employ neural networks to encode the instance semantics into a vector. In this study, we implement the instance encoder with convolutional neural networks (CNNs), given both model performance and time efficiency. 

\textbf{Adversarial Relation Adaptation.}
We follow the standard adversarial transfer learning framework, which looks for a relation discriminator that can distinguish between samples having different relation distributions.  

\textbf{Weighted Relation Adaptation.}
We propose a relation-gate mechanism to identify the unrelated source relations/samples and down-weight their importance automatically to tackle the problem of \emph{negative transfer}.

\subsection{Instance Encoder}

\begin{figure}[!htbp]
  \centering
  \includegraphics[scale=0.4]{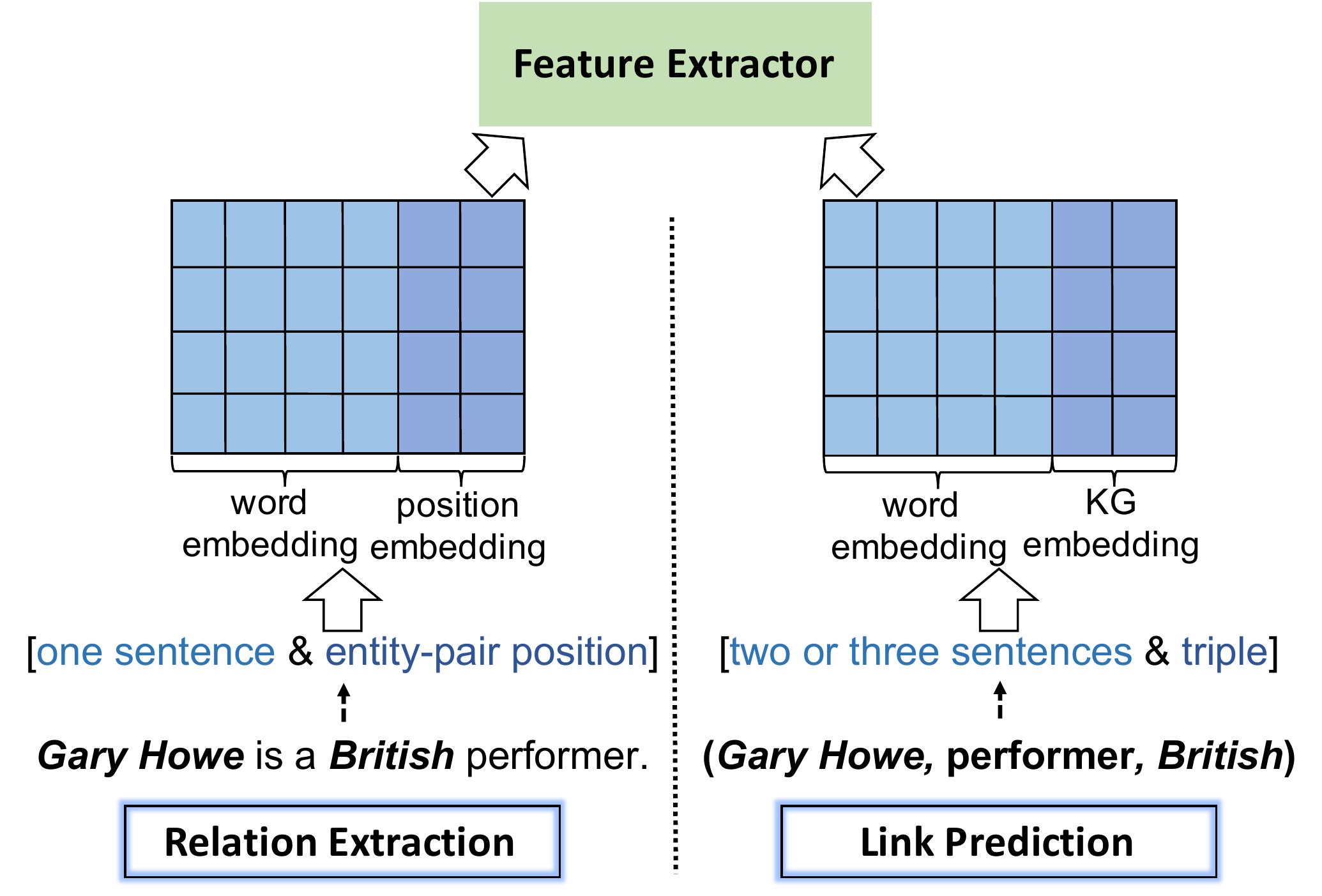}
  \caption{Instance Encoder.}\label{fig:encoder}
\end{figure}

We utilize different encoders for link prediction and relation extraction, as shown in Figure~\ref{fig:encoder}. 

To model the plausibility of a triple, we packed the sentences of $(h, r, t)$ as a single sequence. 
The sentence is the name or description of each entity and relation, e.g., "\emph{Steve Paul Jobs}" or "\emph{Steven Paul Jobs was an American business magnate, entrepreneur and investor.}".
The single input token sequence may contain two sentences of entity $h$ and $r$ or three sentences of $h$, $r$, $t$ packed together. 
To explicitly model the KG structure information, we also utilize pretrained TransE \cite{TransE:conf/nips/BordesUGWY13} embedding to form the final input representation. To extract the relation between entities from corpus, we utilize pretrained word embedding and random initialized position embedding \cite{zeng2014relation}. Given a sentence $s = \{w_1, ..., w_L\}$, where $w_i$ is the $i$-th word in the sentence, the input is a matrix consisting of $L$ vectors $V = [v_1, \cdots, v_L]$.

We apply non-linear transformation $F(\cdot)$ to the vector representation of $V$ to derive a feature vector $\mathbf{f} = F(V; \theta)$. 
We choose two convolutional neural architectures, CNN \cite{zeng2014relation} and PCNN \cite{zeng2015distant}, to encode input embeddings into instance embeddings. Other neural architectures such as RNN \cite{zhang2015relation} and more sophisticated approaches such as ELMo \cite{peters2018deep} and BERT \cite{devlin2018bert} can also be used. 
We adopt the unshared feature extractors for both relations since unshared extractors are able to capture more relation specific features \cite{tzeng2017adversarial}. We  train the source discriminative model $C(F_s(x; \theta))$ for the classification task by learning the parameters of the source feature extractor $F_s$ and classifier $C$:

\begin{equation}
\min \limits_{F_s, C} \mathbb{L}_s =\mathbb{E}_{x, y  \sim p_s(x,y)} L(C(F_s(x; \theta)),y)
\end{equation} 
where $y$ is the label of the source data $x$, $L(\cdot)$ is the loss function for classification. Afterwards, the parameters of $F_s$ and $C$ are fixed. Notice that, it is easy to obtain a pretrained model from the source relations, which is convenient in real scenarios.

\subsection{Adversarial Relation Adaptation}
To built \emph{Relation Adversarial Network}, we follow the standard adversarial transfer learning framework, which is a popular solution in both computer vision \cite{tzeng2017adversarial} and  NLP \cite{shah2018adversarial}. Similarly but differently, the adversarial learning procedure for wRAN is a two-player game, where the first player is the relation discriminator $D$ trained to distinguish the source relation from the target relation, and the second player is the feature extractor $F$ which is trained simultaneously to confuse the relation discriminator and seeks to learn common-features adaptable from source to target relation.
%To learn relation-invariant representations, we utilize the adversarial transfer learning framework, which is a popular solution in both computer vision \cite{tzeng2017adversarial} and  NLP \cite{shah2018adversarial}. 
The general idea is to learn both relation discriminative and relation-invariant features, where the loss of the label predictor of the source data is minimized while the loss of the relation discriminator is maximized. 

\begin{equation}\begin{split}
\min \limits_{F_s,F_t} \max \limits_{D} \mathbb{L}(D,F_s,F_t) =& \mathbb{E}_{x \sim p_s(x)}[\log D(F_s(x))]+ \\&
\mathbb{E}_{x \sim p_t(x)}[1-\log D(F_t(x))]
\end{split}
\end{equation}
where $F_s$ and $F_t$ are the feature extractors for the source and the target data, respectively, and $D$ is a binary relation discriminator (corresponding to the discriminator in the original GAN) with all the source data labeled as 1 and all the target data labeled as 0.

\subsection{Weighted Relation Adaptation}

In practice, it is expected to adopt features from multiple related relations to the target relation. However, different source relations might have different contributions to the transfer. More importantly, those unrelated or outlier relations might cause \emph{negative transfer}. Moreover, the negative transfer might also happen at the sample level, i.e., different samples in the same relation might also have a different effect on the transfer. Considering the relation \emph{educated\_at} as an example,  given an instance \emph{"\textbf{James Alty} graduated from \textbf{Liverpool University}"} from target relation in the relation extraction task, semantically, a   more similar  instance of \emph{"\textbf{Chris Bohjalian} graduated from \textbf{Amherst College}"}  will  provide more  reference  while  a dissimilar  instance \emph{"He was a professor at \textbf{Reed College} where he taught \textbf{Steve Jobs}"} may have little contributions. To address these challenges, we  propose \emph{weighted relation adaptation} to evaluate the significance of each source relation/sample to the target relation  through two perspectives: \textbf{relation correlation}  and \textbf{instance transfer ability}.

\textbf{Relation Correlation.}
Given that not all source relations are beneficial and can be adapted to the target relation, it is intuitive to assign different weights to different relations to lower the negative transfer effect of outlier relations and encourage positive transfer for those more similar or related relations. For example, it is necessary to lower the \emph{capital} relation to mitigating negative transfer. Notice that, the output of the source classifier to each data point provides a probability distribution over the source label, we average the label predictions on all target data from the source classifier as relation weights.  The intuition here is that the probability of assigning the target data to the source outlier relations is sufficiently small. %We average the label predictions on all target data from the source classifier as relation weights \cite{long2018}. 
Practically,  the source classifier $C(F_s(x_i))$  reveals a probability distribution over the source relation space $r_s$. This distribution characterizes well the probability of assigning $x_i$ to each of the $|r_s|$ relations.   We average the label predictions $\hat y_i = C(F_s(x_i)),\ x_i \in D_t, $ on all target data since it is possible that the source classifier can make a few mistakes on some target data and assign large probabilities to false relations or even to outlier relations. The weights indicating the contribution of each source relations to the training can be computed as follows:

\begin{equation}
w^{relation} = \frac{1}{n_t}\sum_{i=1}^{n_t} \hat y_i \label{relation}
\end{equation}
where  $w^{relation}$ is a $|r_s|$-dimensional weight vector quantifying the contribution of each source relation.  

\textbf{Instance Transfer Ability.}
Although the relation weights provide a global weights mechanism to de-emphasize the effect of outlier relations, different instances have different impacts, and not all instances are transferable.  It is necessary to learn fine-grained instance weights to lower the effects of samples that are nontransferable. 

Given the instance encoder of the source and target relations, we utilize a pretrained auxiliary relation discriminator for instance weights learning. We regard the output of the optimal parameters of the auxiliary relation discriminator as instance weights. The concept is that if the activation of the auxiliary relation discriminator is large, the sample can be almost correctly discriminated from the target relation by the discriminator, which means that the sample is likely to be nontransferable  \cite{zhang2018importance}. 

Practically, given the learned $F_s(x)$ from the  instance encoder, a relation  adversarial loss is used to reduce the shift between relations by optimizing $F_t(x)$ and auxiliary relation discriminator $D_a$:

\begin{equation}\begin{split}
\min \limits_{F_t} \max \limits_{D_a} \mathbb{L}_d (D_a,F_s,F_t) =& \mathbb{E}_{x \sim p_s(x)}[\log D_a(F_s(x))]+ \\&
\mathbb{E}_{x \sim p_t(x)}[1-\log D_a(F_t(x))]
\label{eq: firstgan}
\end{split}
\end{equation}

To avoid a degenerate solution, we initialize $F_t$ using the parameter of $F_s$. The auxiliary relation discriminator is given by $D_a(f) = p(y=1|x)$ where x is the input from the source and the target relation.    If  $D_a(f) \equiv 1$, then it is  likely that the sample is nontransferable, because it  can be almost perfectly discriminated from the target distribution by the relation discriminator. The contribution of these samples should be small. Hence, the weight function should be inversely related to $D_a(f)$, and a natural way to define the  weights of the source samples is: 

\begin{equation}
w_i^{instance} = \frac{1}{\frac{D_a(F_s(x))}{D_a(F_t(x))}+1} = 1-D_a(f)
\end{equation}

\textbf{Relation-gate Mechanism.} 
Both relation and instance weights are helpful.  However, it is obvious that the weights of different granularity have different contributions to different target relations.  On the one hand, for target relations (e.g., \emph{located\_in}) with relatively less semantically similar source relations,  it is advantageous to strengthen the relation weights to reduce the negative effects of outlier relations. On the other hand, for target relations (e.g., \emph{educated\_in}) with many semantically similar source relations (e.g., \emph{live\_in},   \emph{was\_born\_in}),  it is difficult to differentiate the impact of different source relations, which indicates the necessity of learning fine-grained instance weights.  

Thus for an instance in the source relation with label $y_j$, the weight of this instance is:

\begin{equation}
w_i^{total}=\alpha w_i^{instance}+(1-\alpha) w_{j}^{relation}
\end{equation}
where  $w_{j}^{relation}$ is the value in the $j$th-dimension of $w^{relation}$.  We normalize the weight $w_i^{total} = \frac{n_sw_i^{total}}{\sum_{i=1}^{n_s}w_i^{total}}$.
$\alpha$  is the output  of relation-gate  to explicitly balance the  instance and relation weights which is  computed as below.

\begin{equation}
\alpha = \sigma(W_rF_t(x))
\end{equation}
where $\sigma$ is the activation function, $W_r$ is the weight matrix. 
\subsection{Initialization and Training}

\textbf{Objective Function.}
The overall objectives of  our approach  are  $\mathbb{L}_s$, $\mathbb{L}_d$ and:

\begin{equation}
\begin{split}
\min \limits_{F_t} \max \limits_{D_r} &  \mathbb{L}_w(C,D_r,F_s,F_t)=\\
& \mathbb{E}_{x \sim p_s(x)}[w^{total}\log D_r(F_s(x))]+  \\
&   \mathbb{E}_{x \sim p_t(x)}[1-\log D_r(F_t(x))]
\end{split}\label{eq: allpro}
\end{equation}
where $D_r$ is the relation adversarial. Note that,  weights $w^{total}$\footnote{The weights can be updated in an iterative fashion when $F_t$ changes. However, we found no improvement in experiments, so we compute the weights and fix them.} are automatically computed and assigned to the source relation data to de-emphasize the outlier relation and nontransferable instances regarding partial adaptation, which can mitigate negative transfer. The overall training procedure is shown below.

%The objectives are optimized in stages. $F_s$ and $C$ are pre-trained on the source domain data and fixed afterward.  Then, the relation weights are learned.  Subsequently, the $D_a$, $D_r$ and $F_t$ are optimized simultaneously without the need to revisit $F_s$ and $C$. Note that $D_a$ is only used for obtaining the instance level weights for the source relations using $F_s$ and current $F_t$, while $D_r$ plays the minimax game with the target relation feature extractor to update $F_t$. We insert a gradient reversal layer (GRL) \cite{ganin2016domain}  to multiply the gradient by -1 for the feature extractor to simultaneously learn the feature extractor $F_t$ and relation classifier $D_a$, $D_r$. The additional computational complexity of our model compared to that of the baselines are attributed to the pretraining of $F_s$  and $C$ to obtain the relation weights and accomplish the pretraining of $F_t$ and $D_a$ to obtain instance weights.  They have a  computation complexity that is linear to the number of training samples.  Moreover, those weights can be shared or as initial values for different datasets (sharing relations) to accelerate training. 

\begin{algorithm}[th]
\caption{Overall Training Procedure}
\label{algorithm: training}

\begin{algorithmic}[1]
\State Pre-train $F_s$ and $C$  on the source domain and fix all parameters afterward.

\State Compute relation weights.

\State Pre-train  $F_t$ and  $D_a$ and compute instance  weights, then fix parameters of $D_a$.

\State Train $F_t$ and $D_r$, update the parameters  of $F_t$  through GRL. 
\end{algorithmic}
\end{algorithm}

\section{Experiments}

\subsection{Datasets and Evaluation}

\textbf{Knowledge Graph.}
Existing benchmarks for KGC, such as FB15k-237 \cite{toutanova2015representing} and YAGO3-10 \cite{mahdisoltani2013yago3} are all small subsets of real-world KGs. These datasets consider the same set of relations during training and testing and often contain sufficient training triples for each relation. To construct datasets for low-resource learning, we go back to original KGs and select those relations that have sparse triples as low-resource relations. We refer to the rest of the relations as source relations since their triples provide important knowledge for us to match entity pairs. Our first dataset FB1.5M is based on Freebase \cite{bollacker2008freebase}, where we remove those inverse relations. We select 30 relations with less than 500 triples as target low-resource relations and the rest as source relations. To show that our model is able to operate on standard benchmarks, we also conduct experiments on FB15k-237. We follow a similar process of filtering low-resource relations to construct FB15k-237-low as a target low-resource dataset. The dataset statistics are shown in Table~\ref{tab:dataset}.

\begin{table}[!htbp]
 \caption{Summary statistics of datasets}
    \label{tab:dataset}
    \centering
    \begin{tabular}{c|ccccc}
    \toprule
    \textbf{Dataset} & \textbf{\#Rel} & \textbf{\#Ent} & \textbf{\#Train} & \textbf{\#Dev} & \textbf{\#Test} \\
    \midrule
     FB1.5M & 4,018 & 1,573,579 & 799,104 & 20,254  & 21,251  \\
     FB15K-237-low & 237 & 14,541 & 272,116 &10,576  &11,251  \\
     ACE05 & 6 & 19,684& 351 & 80 & 80 \\
     Wiki-NYT & 60 &32,614  & 615,691 & 231,345 & 232,549 \\
    \bottomrule
    \end{tabular}
\end{table}

% ===== in order to put the table on the page it first mentioned =====
\begin{table*}[!htbp]
\caption{Main results of entity prediction on the FB1.5M and FB15k-237-low dataset.}
    \label{tab:linkprediction}
   %\fontsize{8}{10}\selectfont
\begin{tabular}{c|ccccc|ccccc}
\toprule
 
\multirow{2}{*}{Model} & \multicolumn{5}{c|}{\textbf{FB1.5M}} & \multicolumn{5}{c}{\textbf{FB15K-237-low}} \\
\cline{2-11}  
& \textbf{MRR} & \textbf{MR} & \textbf{HIT@1} & \textbf{HIT@3} & \textbf{HIT@10} & \textbf{MRR} & \textbf{MR} & \textbf{HIT@1} & \textbf{HIT@3} & \textbf{HIT@10}  \\
 
\midrule 
 TransE \cite{TransE:conf/nips/BordesUGWY13} &0.075& 15932 & 0.046 & 0.087 & 0.128 &0.203& 325 & 0.105 & 0.235 & 0.312 \\   
 TransH \cite{TransH:conf/aaai/WangZFC14} &0.079& 15233 & 0.049 & 0.089 & 0.130 &0.205&352  & 0.110 & 0.206 & 0.322 \\   
 TransR \cite{TransR:conf/aaai/LinLSLZ15} &0.081& 15140 &0.050  & 0.090 & 0.135 &0.209& 311 & 0.095 & 0.265  & 0.342 \\   
 TransD \cite{TransD:conf/acl/JiHXL015}  &0.085& 15152 &0.048  &0.089 & 0.139 &0.212&  294 & 0.112 & 0.283 & 0.351 \\   
 DistMult \cite{DistMult:conf/iclr/2015} &0.065& 21502 & 0.036 & 0.076 & 0.102 &0.185& 411 & 0.123 & 0.215 & 0.276  \\   
 ComplEx \cite{ComplEx:conf/icml/TrouillonWRGB16} &0.072& 16121  &0.041  & 0.090 & 0.115 &0.192& 508 &  0.152 & 0.202 & 0.296  \\ 
 ConvE \cite{ConvE:conf/aaai/DettmersMS018} &-& - & - & - & - &0.224& 249 & 0.102 & 0.262 &0.352 \\ 
 Analogy \cite{ANALOGY:conf/icml/LiuWY17} &0.116& 13793 & 0.102 &0.145 & 0.205  &0.235& 267  & 0.096 & 0.256 & 0.357 \\  
 KBGAN \cite{cai2017kbgan} &-& - & - & - & - & - & - & 0.142 & 0.283 & 0.342 \\  
 RotatE \cite{RotatE:Sun2019RotatE}  &0.125& 10205 & 0.135 & 0.253 &0.281  &0.246& 225  & 0.153 & 0.302 & 0.392 \\ 
 KG-BERT \cite{KG-BERT:Yao2019KG}            &0.119& 13549 & 0.093 &0.195  & 0.206 &0.258& 159  & 0.146 & 0.286 & 0.371 \\   
  IterE \cite{IterE:conf/www/ZhangPWCZZBC19} &0.109& 14169 & 0.073 &0.175  & 0.186 &0.238& 178  & 0.152 & 0.275 & 0.361 \\   
\midrule

\textbf{wRAN} & \textbf{0.141}&\textbf{9998} & \textbf{0.199} & \textbf{0.284} & \textbf{0.302} &\textbf{0.262} &\textbf{139} & \textbf{0.212} & \textbf{0.354} & \textbf{0.415} \\   

\bottomrule
\end{tabular}
\end{table*}

\textbf{Text Corpus.}
We use the ACE05\footnote{https://catalog.ldc.upenn.edu/LDC2006T06} dataset to evaluate our approach by dividing the articles from six genres and seven relations into respective domains: broadcast conversation (bc), broadcast news (bn), telephone conversation (cts), newswire (nw), usenet (un) and weblogs (wl).  We use the same data split followed by \cite{fu2017domain}, in which bn $\&$ nw are used as the source, half of bc, cts, and wl are used as the target for training (no label available in the unsupervised setting), and the other half of bc, cts, and wl are used as target for test. We split 10\% of the training set to form the development set and fine-tune hyper-parameters.  We calculate the  proxy A-distance \cite{mansour2009domain} $d_{\mathcal{A}}=2(1-2 \epsilon)$, where $\epsilon$     is the generalization error of a
trained domain classifier. The average A-distance from all sources to target domains is about 1.1, which shows there exists a big difference between domains.  
We conducted two kinds of experiments. 
The first is standard adaptation, in which the source and target have the same relation set. 
The second is partial adaptation, in which the target has only half of the source relations. 
For the DS setting, we utilize two existing datasets NYT-Wikidata \cite{zeng2016incorporating}, which align Wikidata \cite{vrandevcic2014wikidata} with New York Times corpus (NYT), and Wikipedia-Wikidata \cite{sorokin2017context}, which align Wikidata with Wikipedia. We filter 60 shared relations to construct a new dataset Wiki-NYT, in which Wikipedia contains source relations and the NYT corpus contains target relations. We conducted partial DA experiments (60 relations $\rightarrow$ 30 relations), and randomly choose half of the relations to sample the target data.
Our datasets are available at \url{https://github.com/zxlzr/RAN}.

\subsection{Parameter Settings}
To fairly compare the results of our models with those baselines, we set most of the experimental parameters following \cite{fu2017domain,lin2016neural}. 
We train GloVe \cite{pennington2014glove} word embeddings on the Wikipedia and  NYT corpus with 300 dimensions. In both the training and testing set, we truncate sentences with more than 120 words into 120 words.  To avoid noise at the early stage of training, we use a similar schedule method as \cite{ganin2014unsupervised} for the trade-off parameter to update $F_t$ by initializing it at 0 and gradually increasing to a pre-defined upper bound. The schedule is defined as: $\Phi = \frac {2 \cdot u} {1+ exp(-\alpha \cdot p)} - u$, where $p$ is the training progress linearly changing from 0 to 1, $\alpha$ = 1, and $u$ is the upper bound set to 0.1 in our experiments. 
% As data imbalance will severely affect the model training and cause the model only sensitive to relations with a high proportion, we cut the propagation of loss from \emph{NA}  to relieve the impact of \emph{NA} followed by \cite{ye2017jointly}, and we set its loss as 0. We implement our approach via Tensorflow \cite{abadi2016tensorflow} and adopt the Adam optimizer  \cite{kingma2014adam}.
We implemented baselines via OpenKE\footnote{\url{https://github.com/thunlp/OpenKE}} and OpenNRE\footnote{\url{https://github.com/thunlp/OpenNRE}}.

\subsection{Link Prediction}
The link (entity) prediction task predicts the head entity $h$ given $(?, r, t)$ or predicts the tail entity $t$ given $(h,r,?)$ where $?$ means the missing element. The results are evaluated using a ranking produced by the scoring function $f(h, r, t)$  on test triples. Each correct test triple $(h, r, t)$ is corrupted by replacing either its head or tail entity with every entity $e \in \mathbb{E}$; then these candidates are ranked in descending order of their plausibility head or tail entity with every entity $e \in \mathbb{E}$, then these candidates are ranked in descending order of their plausibility score. We report three common metrics, Mean Reciprocal Rank (MRR) and  Mean Rank (MR) of correct entities and Hits@N which means the proportion of correct entities in top $N$. A lower MR is better while a higher MRR and Hits@N is better.

We compare wRAN  with multiple state-of-the-art KG embedding methods as follows: \textbf{TransE} \cite{TransE:conf/nips/BordesUGWY13} and its extensions \textbf{TransH} \cite{TransH:conf/aaai/WangZFC14}, \textbf{TransD} \cite{TransD:conf/acl/JiHXL015}, \textbf{TransR} \cite{TransR:conf/aaai/LinLSLZ15}; \textbf{DistMult} \cite{DistMult:conf/iclr/2015} is a generalized  framework  where entities are low-dimensional vectors learned from a neural network and relations are bilinear and/or linear mapping functions; \textbf{ComplEx} \cite{ComplEx:conf/icml/TrouillonWRGB16} extends DistMult into the complex number field; \textbf{ConvE} \cite{ConvE:conf/aaai/DettmersMS018} is a multi-layer convolutional network model for link prediction;  \textbf{Anology} \cite{ANALOGY:conf/icml/LiuWY17} is a framework which  optimize the latent representations with respect to the analogical properties of the embedded entities and relations; \textbf{KBGAN} \cite{cai2017kbgan} is an adversarial learning framework for generating better negative examples to train knowledge graph embedding models; \textbf{RotaE} \cite{RotatE:Sun2019RotatE} defines each relation as a rotation from the source entity to the target entity in the complex vector space; \textbf{KG-BERT} \cite{KG-BERT:Yao2019KG} takes entity and relation descriptions of a triple as input and computes scoring function of the triple with BERT. 

Following \cite{ConvE:conf/aaai/DettmersMS018} , we only report results under the filtered setting which removes all corrupted triples that appeared in training, development, and test set before getting the ranking lists. Table~\ref{tab:linkprediction} shows link prediction performance of various models.   We can observe that: 
1) \textbf{wRAN} can achieve lower MR than baseline models, and it achieves the lowest mean ranks on FB1.5M and FB15k-237-low to our knowledge. 
2) The Hits@N scores of \textbf{wRAN} are higher than in state-of-the-art methods. \textbf{wRAN} can leverage the relation-invariant features from source relations to benefit the low-resource target relations. 
3)  The improvement of \textbf{wRAN} over baselines on FB1.5M is larger than FB15K-237-low because  FB1.5M has more source relations, which can leverage more knowledge from semantically related relations.

\begin{table}[!htbp]
\caption{Triple classification accuracy (in percentage)  on the FB1.5M and FB15K-237-low dataset.}
    \label{tab:tripleclassification}
   %\fontsize{8}{10}\selectfont
\begin{tabular}{c|cc|c}
\toprule

 \textbf{Method} & \textbf{FB1.5M} & \textbf{FB15K-237-low} & \textbf{Avg.} \\
 \midrule
  TransE \cite{TransE:conf/nips/BordesUGWY13} & 56.6 & 74.2 & 65.4 \\
  TransH \cite{TransH:conf/aaai/WangZFC14} & 56.9 & 75.2 & 66.1 \\
  TransR \cite{TransR:conf/aaai/LinLSLZ15} &5 6.4 & 76.1 & 66.3 \\
  TransD \cite{TransD:conf/acl/JiHXL015} & 60.2 & 76.5 & 68.4 \\
  DistMult \cite{DistMult:conf/iclr/2015} & 60.5 & 82.2 & 71.4 \\
  ComplEx \cite{ComplEx:conf/icml/TrouillonWRGB16} & 61.2 & 83.3 & 72.3 \\
  Analogy \cite{ANALOGY:conf/icml/LiuWY17} & 63.1 & 85.2 & 74.2 \\
  ConvE \cite{ConvE:conf/aaai/DettmersMS018} & 63.0 & 84.2 & 73.6 \\
  KG-BERT \cite{KG-BERT:Yao2019KG} & 67.2 & 87.5 & 77.4\\
  IterE \cite{IterE:conf/www/ZhangPWCZZBC19} & 65.2 & 86.4 & 75.8\\
\midrule
\textbf{wRAN} & \textbf{69.2} & \textbf{90.9} & \textbf{80.0} \\
  
\bottomrule
\end{tabular}

\end{table}

Triple classification aims to judge whether a given triple $(h, r, t)$ is correct or not.
Table~\ref{tab:tripleclassification} presents triple classification accuracy of different methods on FB1.5M and FB15K-237-low. We can see that \textbf{wRAN} clearly outperforms all baselines by a large margin, which shows the effectiveness of our method. We ran our models ten times and found the standard deviations are less than 0.2, and the improvements are significant $(p < 0.01)$. To our knowledge, \textbf{wRAN} achieves the best results so far. For more in-depth performance analysis, we note that TransE could not achieve high accuracy scores because it could not deal with 1-to-N, N-to-1, and N-to-N relations. TransH, TransR,  TransD outperform TransE by introducing relation to specific parameters. ConvE shows decent results, which suggests that CNN models can capture global interactions among the entity and relation embeddings.  However, their improvements are still limited to those low-resource relations.

\subsection{Relation Extraction}

\textbf{Evaluation Results on ACE05.}
To evaluate the performance of our proposed  approach, we compared our model with various conventional domain adaptation  models: \textbf{wRAN} is our approach,  \textbf{CNN+DANN} is an unsupervised adversarial DA method \cite{fu2017domain},  \textbf{Hybrid} is a composition model that combines  traditional feature-based method, CNN and RNN \cite{nguyen2015combining}, and \textbf{FCM} is a compositional embedding model.  The first experiment takes the standard adaptation setting where we assume the source and target share the same relation set. In this setting, as the evaluation results are shown in Table~\ref{tab:normalace}, we observe that our model achieves performance comparable to that of CNN+DANN, which is a state-of-the-art model, and significantly outperforms the vanilla models without adversarial learning. This shows that adversarial learning is effective for learning domain-invariant features to boost performance as conventional adversarial domain adaptation models. The second experiment takes the partial adaptation setting where we only choose a part of the relations as target and study the weighted adaptation from a larger set of source relations. As shown in Table~\ref{tab:normalace} again, our model significantly outperforms the plain adversarial adaptation model, CNN+DANN, in the partial setting. This demonstrates the efficacy of our hybrid weights mechanism\footnote{Since the adversarial adaptation method significantly outperforms traditional methods \cite{fu2017domain}, we skip the performance comparison with FCM and Hybrid for partial adaptation.}. 

\begin{table}[!htbp]
\caption{F1 score of normal and partial DA on ACE05 dataset. * indicates $p_{value} < 0.01$ for a paired t-test evaluation. }
\label{tab:normalace}
\centering
\begin{small}
\begin{tabular}{c|c|c|c|c}
% \hline
\toprule
\centering \textbf{Standard Adaptation} & \centering \textbf{bc} & \centering \textbf{wl} & \centering \textbf{cts} & \textbf{Avg.} \\
% \hline
\midrule
\centering FCM & 61.90 & N/A & N/A & N/A \\
\centering Hybrid & 63.26 & N/A & N/A & N/A \\
\centering CNN+DANN & 65.16 & 55.55 & 57.19 & 59.30 \\
\hline
\centering \textbf{wRAN} & \textbf{66.15*} & \textbf{56.56*} & 56.10 & \textbf{59.60*} \\
% \hline
% \hline
\bottomrule
\toprule
\centering \textbf{Partial Adaptation} & \centering \textbf{bc}& \centering \textbf{wl} & \centering \textbf{cts} & \textbf{Avg.} \\
% \hline
\midrule
\centering CNN+DANN & 63.17 & 53.55 & 53.32 & 56.68 \\
\hline
\centering \textbf{wRAN} & \textbf{65.32*} & \textbf{55.53*} & \textbf{54.52*} & \textbf{58.92*} \\
% \hline
\bottomrule
\end{tabular}

\end{small}
\end{table}

\textbf{Evaluation Results on Wiki-NYT.}
For the DS setting, we consider the setting of 
(1) \textbf{unsupervised adaptation} in which the target labels are removed, 
(2) \textbf{supervised adaptation} in which the target labels are retained to fine-tune our model. 

\begin{figure}[!htbp]

\label{tab:unsupervise}
  \centering
  \includegraphics[width=0.35\textwidth]{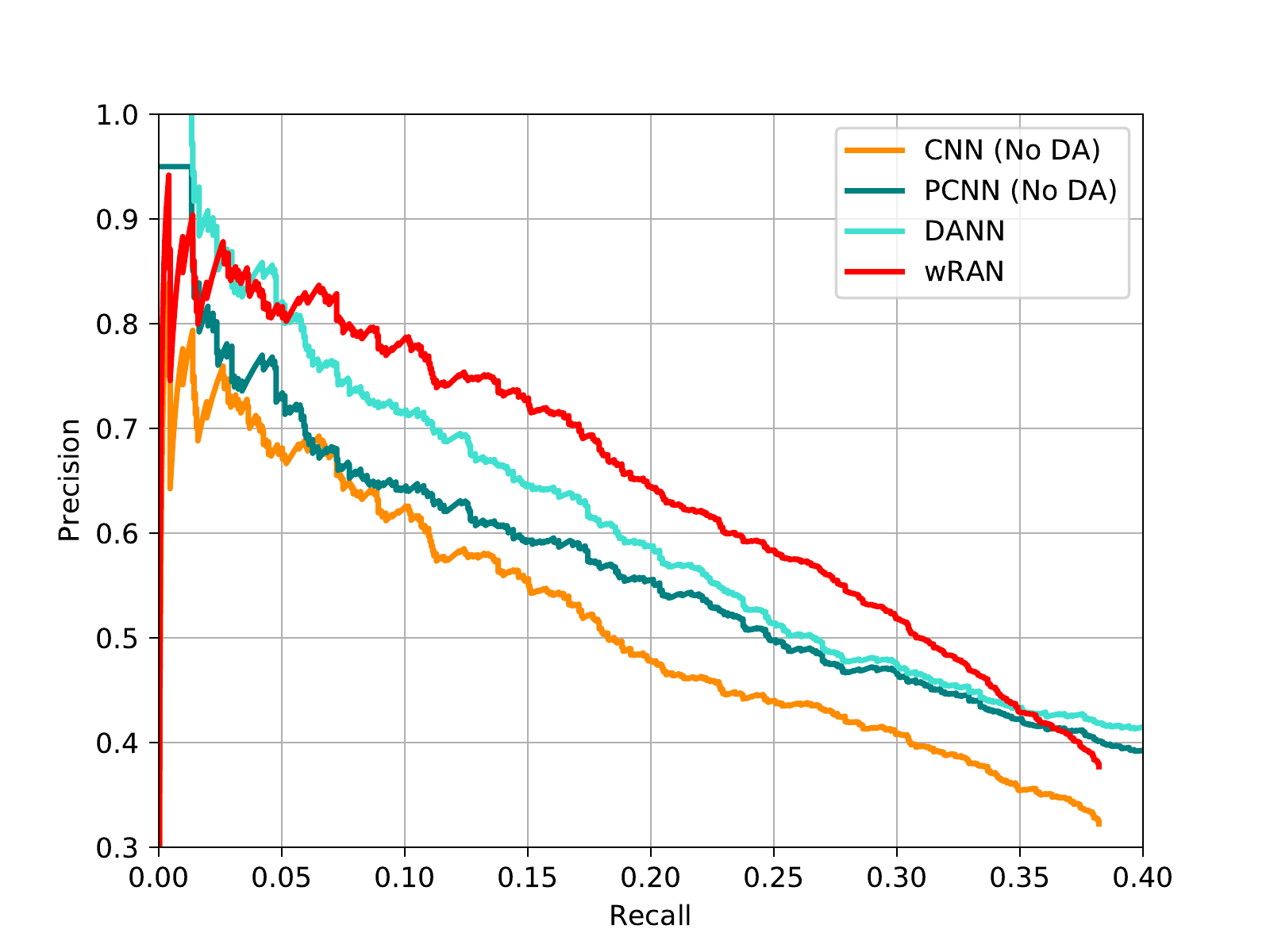}
  \vspace{-3mm}
  \caption{Unsupervised adaptation results.}\label{fig:wikinytpr}
  % \vspace{-5mm}
\end{figure}

\textbf{Unsupervised Adaptation.}
Target relations are unnecessary in unsupervised adaptation. We report the results of our approach and various baselines: \textbf{wRAN} is our unsupervised adaptation approach, \textbf{PCNN (No DA)} and \textbf{CNN (No DA)} are methods trained on the source by PCNN \cite{lin2016neural} and CNN  \cite{zeng2014relation} and tested on the target. Following \cite{lin2016neural}, we perform both held-out evaluation as the \emph{precision-recall curves} shown in Figure~\ref{fig:wikinytpr} and manual evaluation of \emph{top-500 prediction results}, as shown in Table~\ref{tab:unsupervise}.

\begin{table}[!htbp]    
\caption{Precision values of the top 100, 200 and 500 sentences for unsupervised and supervised adaptation. * indicates $p_{value} < 0.01$ for a paired t-test  evaluation.
}
\label{tab:unsupervise}
\renewcommand\tabcolsep{4pt}
\centering
\begin{small}
\begin{tabular}{c|cccc}
\toprule
\centering \textbf{Precision} & \textbf{Top 100} & \textbf{Top 200} & \textbf{Top 500} & \textbf{Avg.} \\
\midrule
\centering CNN (No DA) & 0.62 & 0.60 & 0.59 & 0.60 \\
\centering PCNN (No DA)& 0.66 & 0.63 & 0.61 & 0.63 \\
\centering CNN+DANN & 0.80 & 0.75 & 0.67 & 0.74 \\
\hline
\centering CNN & 0.85 & 0.80 & 0.69 & 0.78 \\
\centering PCNN & 0.87 & 0.84 & 0.74 & 0.81 \\
\centering Rank+ExATT &  0.89 & 0.84 & 0.73 & 0.82 \\
\hline
\centering wRAN & \textbf{0.85*} & \textbf{0.83*} & \textbf{0.73*} & \textbf{0.80*} \\
\centering +25\% & 0.88 & 0.84 & 0.75 & 0.82 \\
\centering +50\% & 0.89 & 0.85  & 0.76 & 0.82 \\
\centering +75\% & 0.90 & 0.85  & 0.77 & 0.83 \\
\centering +100\% & 0.88 & 0.86 & 0.77 & 0.83 \\
\bottomrule
\end{tabular}

% \vspace{-5mm}
\end{small}
\end{table}

We observe that 
(1) Our approach achieves the best performance among all the other unsupervised adaptation models, including CNN+DANN. This further demonstrates the effectiveness of the hybrid weights mechanism. 
(2) Our unsupervised adaptation model achieves nearly the same performance even with the supervised approach CNN; however, it does not outperform PCNN. This setting could be advantageous as in many practical applications, the knowledge bases in a vertical target domain may not exist at all or must be built from scratch.

\textbf{Supervised Adaptation.}
Supervised Adaptation does require labeled target data; however, the target labels might be few or noisy. In this setting, we fine-tune our model with target labels. We report the results of our approach and various baselines: \textbf{wRAN+$k\%$} implies fine-tuning our model using $k\%$ of the target domain data, \textbf{PCNN} and \textbf{CNN} are the methods trained on the target domain by PCNN \cite{lin2016neural} and CNN \cite{zeng2014relation}, and \textbf{Rank+ExATT} is the method trained on the target domain which integrates PCNN with a pairwise ranking framework \cite{ye2017jointly}.

\begin{figure}[!htbp]
  \centering
  \includegraphics[width=0.35\textwidth]{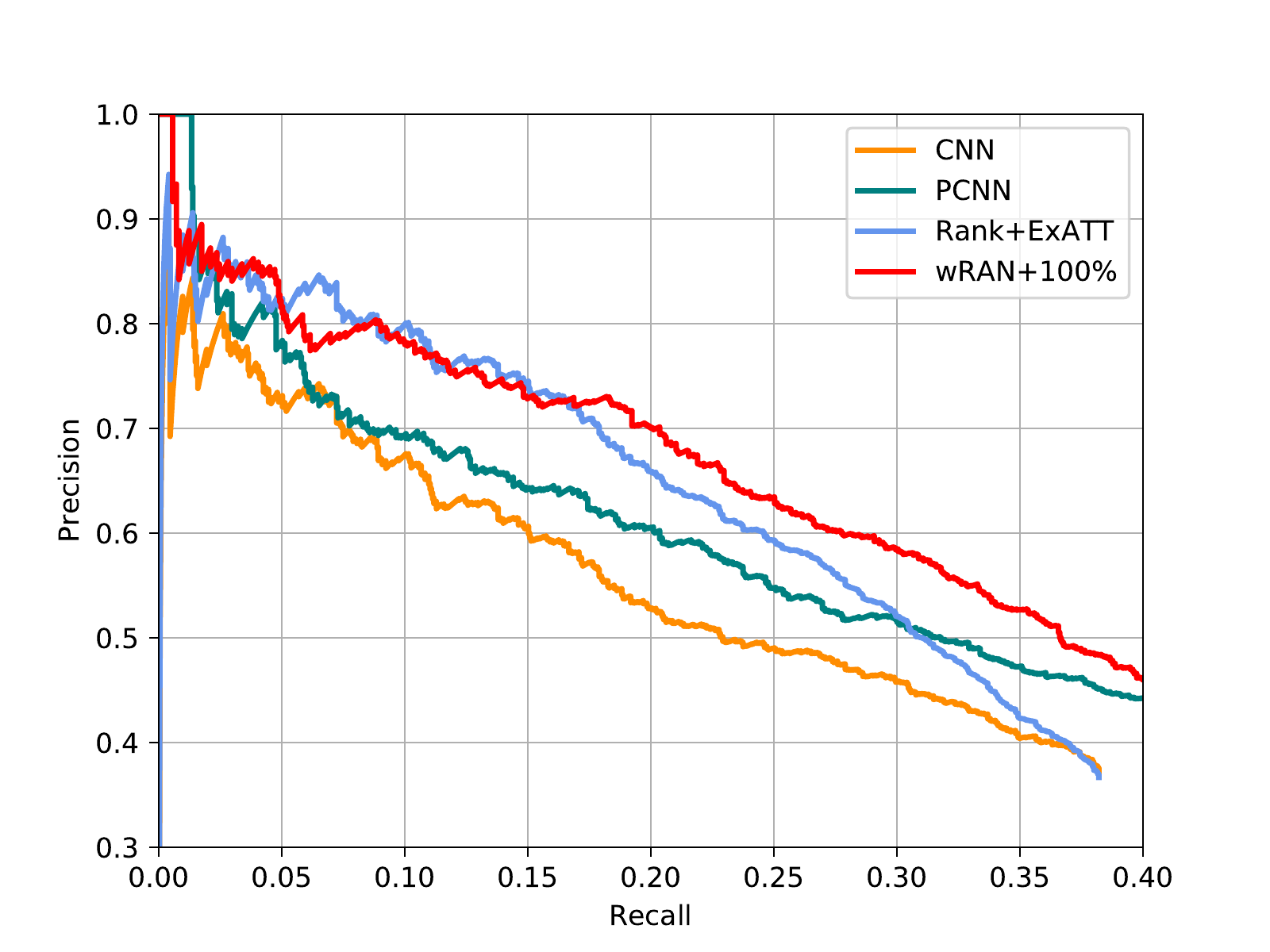}
  \vspace{-3mm}
  \caption{Supervised adaptation results.}\label{fig:pretrain}
\end{figure}

As shown in  Figure~\ref{fig:pretrain} and Table~\ref{tab:unsupervise}, we observe that: (1) Our fine-tuned model +100\% outperforms both CNN and PCNN and achieves results comparable to that of Rank+ExATT. 
%The case study results in Table~\ref{tab:case3} further shows that our model can correct noisy labels to some extent due to the relatively high quality of source domain data. 
(2) The extent of improvement from using 0\% to 25\% of target training data is consistently more significant than others such as using 25\% to 50\%, and fine-tuned model with only thousands labeled samples (+25\%) matches the performance of training from scratch with 10$\times$  more data, clearly demonstrating the benefit of our approach. (3) The top 100 precision of fine-tuned model degrades from 75\% to 100\%. This indicates that there exits noisy data which contradict with the data from the source domain. We will address this by adopting additional denoising mechanisms like reinforcement learning, which will be part of our future work.

\subsection{Ablation Study}

To better demonstrate the performance of different strategies in our model, we  separately remove the relation and instance weights. The experimental results of relation extraction and link prediction are summarized in Table~\ref{tab:ablationwikinyt}, Table~\ref{tab:ablationtripleclassification}, and Table~\ref{tab:ablationentityprediction}, respectively. 
\textbf{wRAN} is our method; \textbf{w/o gate} is the method without relation-gate ($\alpha$  is fixed.); 
\textbf{w/o relation} is the method without relation weights ($\alpha=1$); 
\textbf{w/o instance} is the method without instance weights ($\alpha=0$); 
\textbf{w/o both} is the method without both weights.

\begin{table}[!htbp]    
\caption{Ablation study results for relation extraction on Wiki-NYT dataset.}
\label{tab:ablationwikinyt}
\renewcommand\tabcolsep{4pt}
\centering
\begin{small}
\begin{tabular}{c|cccc}
\toprule
\centering \textbf{Precision} & \textbf{Top 100} & \textbf{Top 200} & \textbf{Top 500} & \textbf{Avg.} \\
\midrule
\centering \textbf{wRAN} & \textbf{0.85} & \textbf{0.83} & \textbf{0.73} & \textbf{0.80} \\ 
\midrule
\centering w/o gate & 0.85 & 0.79 & 0.70 & 0.78 \\
\centering w/o relation & 0.81 & 0.76 & 0.66 & 0.74 \\
\centering w/o instance & 0.84 & 0.78 & 0.69 & 0.77 \\
\centering w/o both & 0.80 & 0.75 & 0.65 & 0.73 \\
\bottomrule
\end{tabular}

\vspace{-5mm}
\end{small}
\end{table}

\begin{table}[!htbp]
\caption{Ablation study results of triple classification on the FB1.5M and FB15K-237-low dataset.}
\label{tab:ablationtripleclassification}
   %\fontsize{8}{10}\selectfont
\begin{tabular}{c|cc|c}
\toprule
\textbf{Method} & \textbf{FB1.5M} & \textbf{FB15K-237-low} & \textbf{Avg.} \\
\midrule
\textbf{wRAN} & \textbf{69.2} & \textbf{90.9} & \textbf{80.0}\\
\midrule
  w/o gate & 67.2 & 89.5 & 78.3\\
  w/o relation & 66.0 & 88.3 & 77.2 \\
  w/o instance & 65.9 & 87.6 & 76.7 \\
  w/o both & 65.2 & 87.2 & 76.2 \\
\bottomrule
\end{tabular}

\vspace{-5mm}
\end{table}

\begin{table*}[!htbp]
\caption{Ablation study results of entity prediction on the FB1.5M and FB15k-237-low dataset.}
\label{tab:ablationentityprediction}
   %\fontsize{8}{10}\selectfont
\begin{tabular}{c|ccccc|ccccc}
\toprule
\multirow{2}{*}{Model} & \multicolumn{5}{c|}{\textbf{FB1.5M}} &\multicolumn{5}{c}{\textbf{FB15K-237-low}} \\
\cline{2-11}  
& \textbf{MRR} & \textbf{MR} & \textbf{HIT@1} & \textbf{HIT@3} & \textbf{HIT@10} 
& \textbf{MRR} & \textbf{MR} & \textbf{HIT@1} & \textbf{HIT@3} & \textbf{HIT@10}  \\
\midrule
\textbf{wRAN} & \textbf{0.141}&\textbf{9998} & \textbf{0.199} & \textbf{0.284} & \textbf{0.302} &\textbf{0.262} &\textbf{139} & \textbf{0.212} & \textbf{0.354} & \textbf{0.415} \\  

\midrule
w/o gate & 0.130 & 11020 & 0.155 & 0.286 &0.291 & 0.252 & 155 & 0.192 & 0.343 & 0.399 \\   
w/o relation & 0.126 & 11560 & 0.134 & 0.275 & 0.282 & 0.243 & 181 & 0.175 & 0.340 & 0.382 \\   
w/o instance & 0.120 & 11900 & 0.126 & 0.270 & 0.278 & 0.238 & 202 & 0.171 & 0.332 & 0.376 \\   
w/o both & 0.113 & 12520 & 0.100 & 0.263 & 0.253 & 0.229 & 254 & 0.166 & 0.326 & 0.364 \\ 
\bottomrule
\end{tabular}

% \vspace{-3mm}
\end{table*}

% ===== in order to put the table on the page it first mentioned =====
\begin{table*}[!htbp]
\caption{Examples for Case 1, 2 and 3, $w^{c}$ and $w^{i}$ denote relation and instance weights, respectively.}
\label{tab:case123}
\centering
\fontsize{8.5}{10}\selectfont
    %\begin{tabular}{|p{3.7cm}|p{0.9cm}|p{0.3cm}|p{0.3cm}|p{0.3cm}|}
\begin{tabular}{c|c|c|c|c}
\toprule
\centering \textbf{Triple Instances} & \textbf{Relations} & \textbf{$w^{r}$} & \textbf{$w^{i}$} & \textbf{$\alpha$} \\
\midrule
(\emph{Rio\_de\_Janeiro}, capital\_of, \emph{Brazil})
& capital\_of & 0.2 & 0.3 & 0.2   \\
\hline
(\emph{Malini\_22\_Palayamkottai}, director, \emph{Sripriya}) 
& director & 0.8 & 0.9 & 0.3  \\
\hline
(\emph{George\_Stevens}, director, \emph{The\_Nitwits}) 
& director & 0.6 & 0.6 & 0.5 \\
\hline
(\emph{Camp}, director, \emph{Todd\_Graff}).
& director & 0.5 & 0.4 & 0.5 \\ 
\hline
(\emph{Chris Bohjalian}, edu\_in, \emph{Amherst College}) 
& educated\_in & 0.4 & 0.7 & 0.9 \\
\bottomrule
\toprule
\centering \textbf{Sentence Instances} & \textbf{Relations} & \textbf{$w^{r}$} & \textbf{$w^{i}$} & \textbf{$\alpha$} \\
\midrule
He was born in \emph{Rio\_de\_Janeiro}, \emph{Brazil} to a German father and a Panama nian mother.
& capital\_of & 0.1 & 0.2 & 0.1   \\
\hline
In 2014, he made his Tamil film debut in \emph{Malini\_22\_Palayamkottai} directed by \emph{Sripriya}.
& director & 0.7 & 0.8 & 0.4  \\
\hline
Sandrich was replaced by \emph{George\_Stevens} for the teams 1935 film \emph{The\_Nitwits}.
& director & 0.7 & 0.5 & 0.5 \\
\hline
\emph{Camp} is a 2003 independent musical\_film written and directed by \emph{Todd\_Graff}.
& director & 0.7 & 0.3 & 0.4 \\ 
\hline
\emph{Chris Bohjalian} graduated from \emph{Amherst College}
& educated\_in & 0.4 & 0.7 & 0.9 \\
\bottomrule
\end{tabular}

\end{table*}

We observe that:
(1) Performance significantly degrades when we remove "relation-gate." This is reasonable because the relation category and instances play different roles for different relations, while \textbf{w/o gate} treats weights equally which hurts performance. 
(2) Performance degrades when we remove "relation weights" or "instance weights." This is reasonable because different weights have different effects on de-emphasizing outlier relations or instances.

\subsection{Parameter Analysis}

\begin{figure}[!htbp]
\centering
\subfigure[Link Prediction] { 
  \includegraphics[scale=0.28]{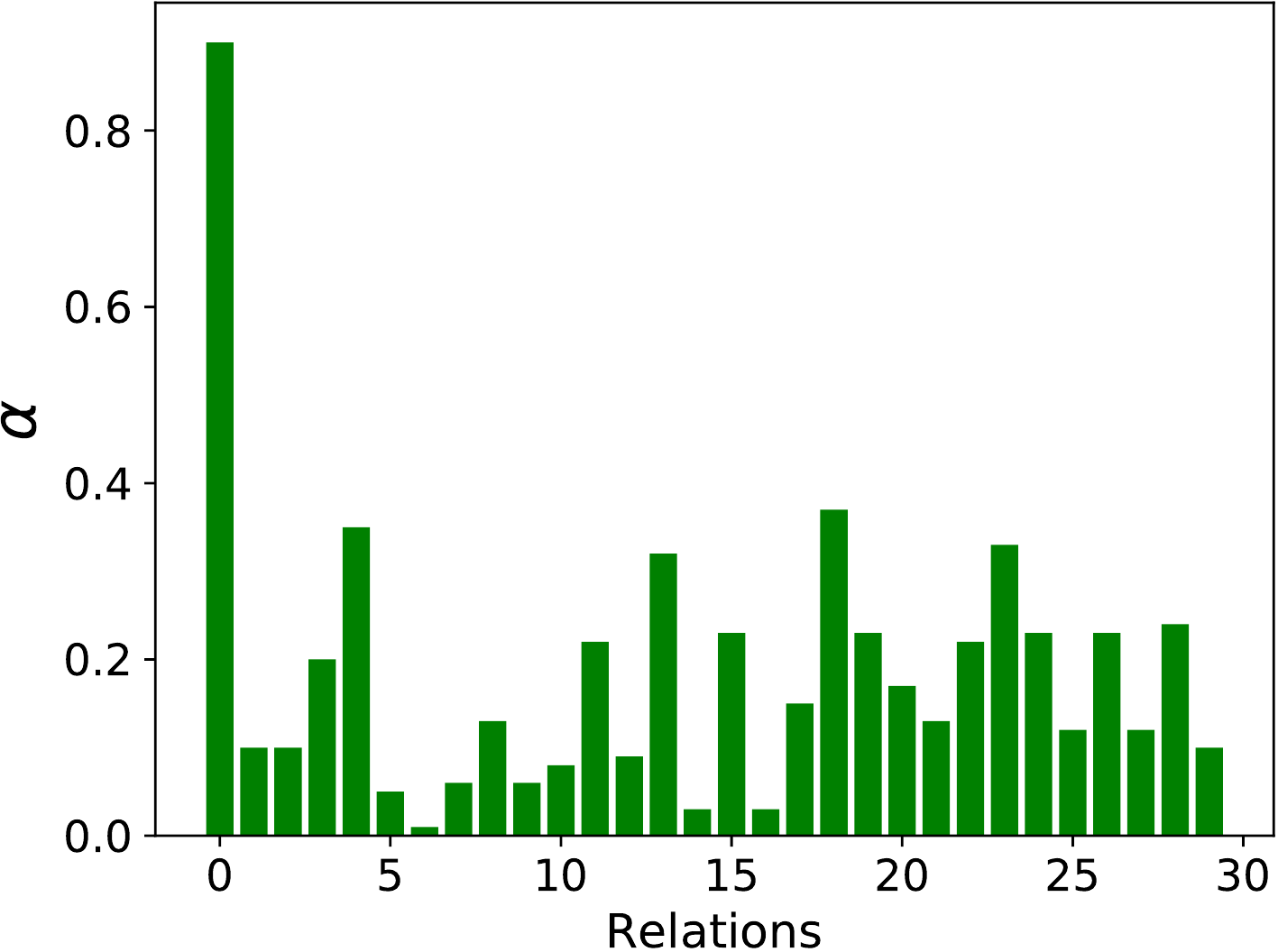}
}
% width=0.22\textwidth
\subfigure[Relation Extraction] { 
\includegraphics[scale=0.28]{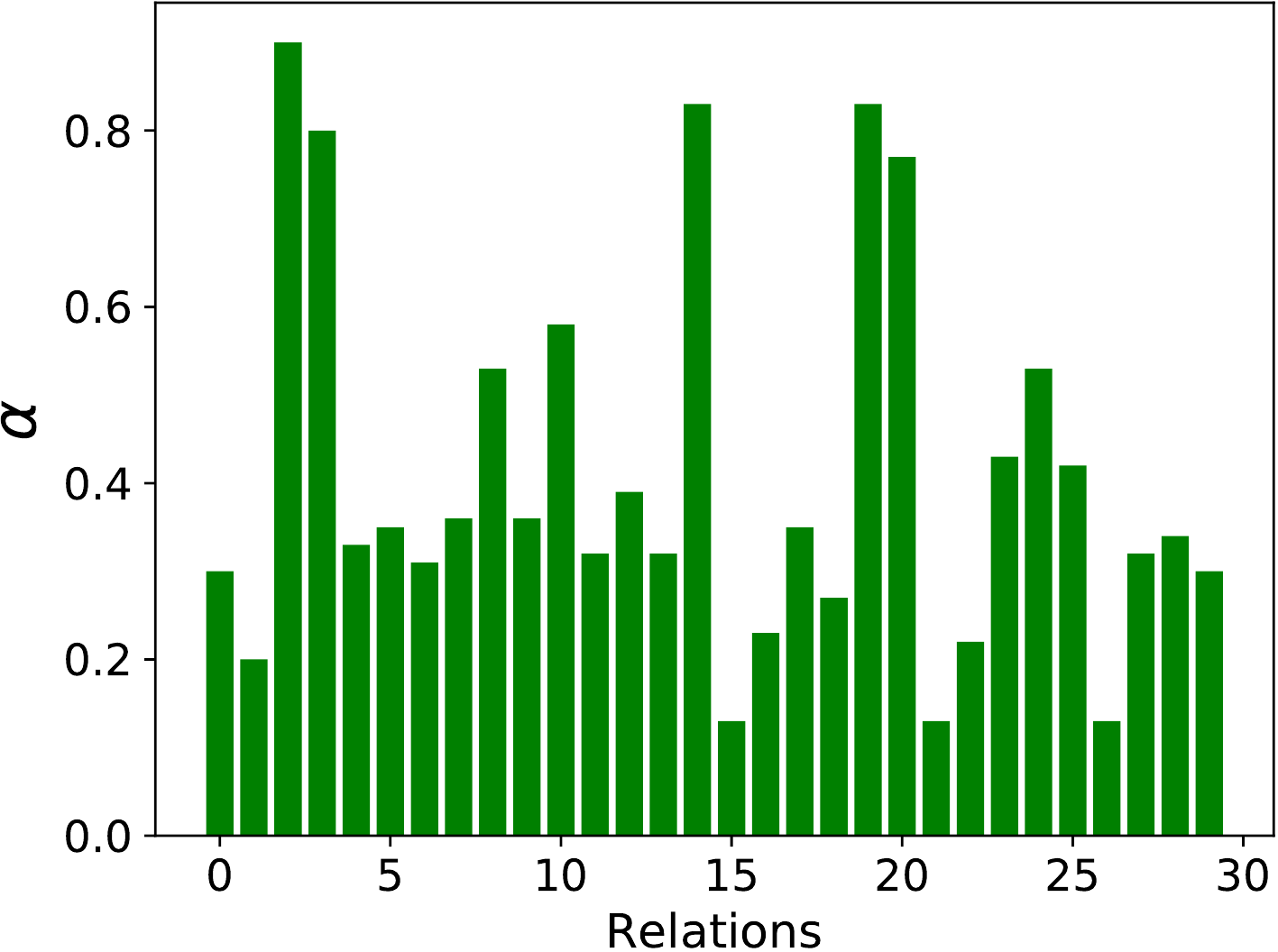}
}
\vspace{-5mm}
\caption{$\alpha$ w.r.t \#Relations.}
\label{fig:para1}
\end{figure}

\textbf{Relation-Gate.} To further explore the effects of relation-gate, we visualize $\alpha$ for all target relations on both FB1.5M and Wiki-NYT dataset.  From the results shown in Figure~\ref{fig:para1}, we observe the following: 
(1) The instance and relation weights have different influences on performance for different relations. Our relation-gate mechanism is powerful to find that instance weights is more important for those relation (e.g., \emph{educated\_at}, \emph{live\_in}
a.k.a., relations with highest $\alpha$  score in Figure~\ref{fig:para1} (b)) while relation weights are more useful for other relations.
(2) The relation weights have relatively more influence on the performance than instance weights for most of the relations due to the noise and variations in instances; however, the relation weights are averaged on all target data and thus less noisy. 

\begin{figure}[!htbp]
\centering
\subfigure[Link Prediction] { 
  \includegraphics[scale=0.23]{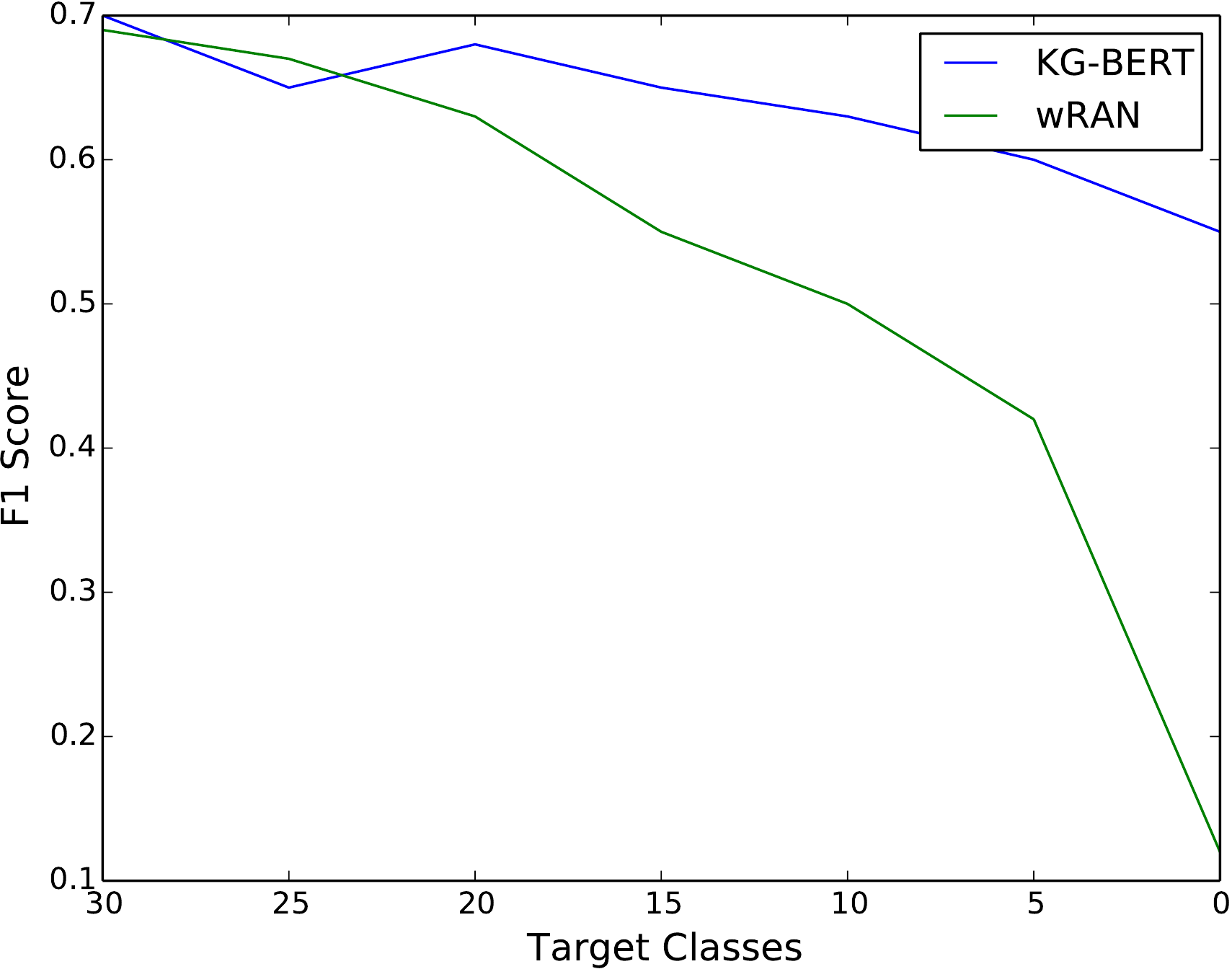}
}
\subfigure[Relation Extraction] { 
\includegraphics[scale=0.23]{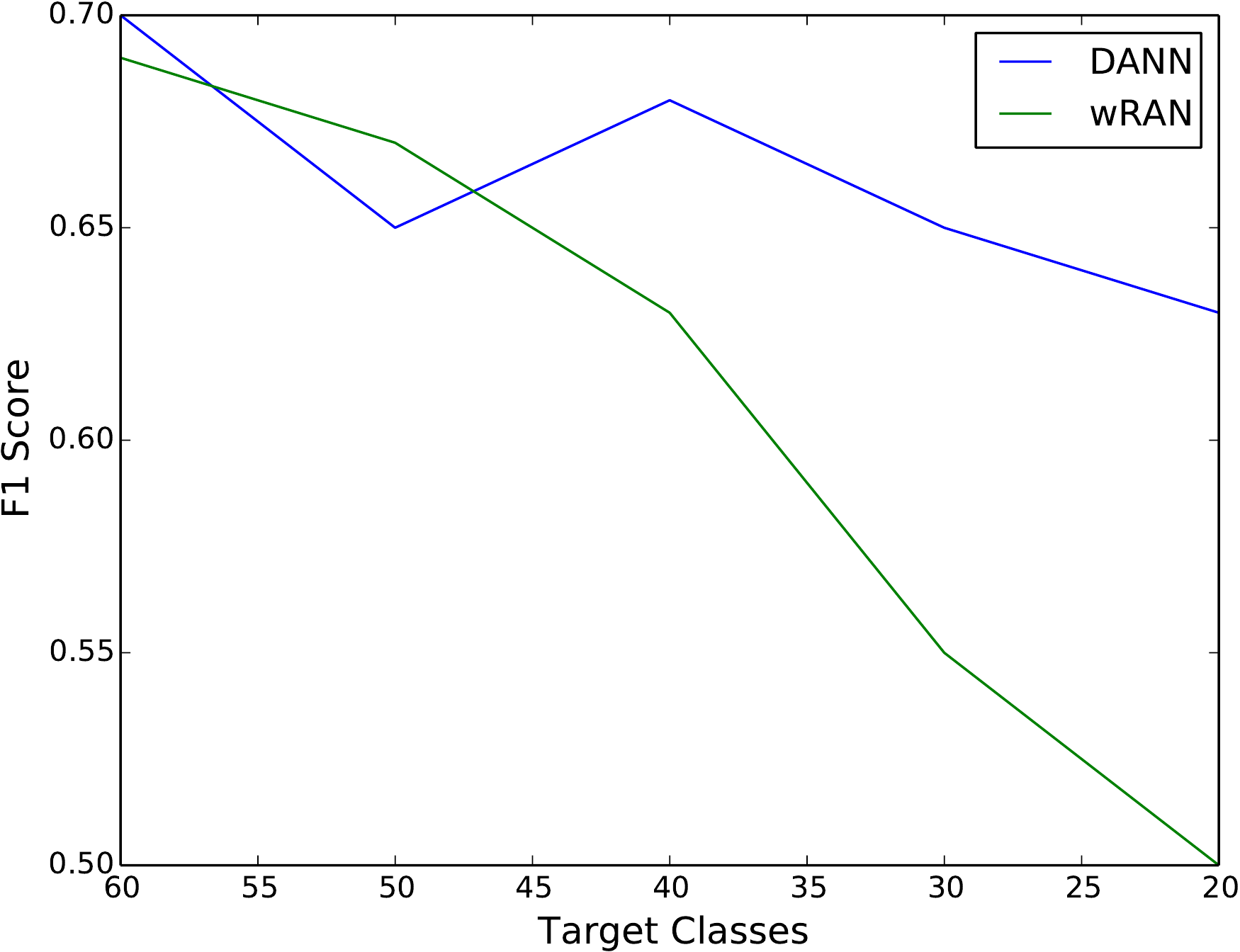}
}
\vspace{-5mm}
\caption{F1 w.r.t \#Target Relations.}
\label{fig:para2}
\end{figure}

\textbf{Different Number of Target Relations.} 
We investigated the transferability of link prediction\footnote{ We choose 60 source relations randomly.} (triple classification) and relation extraction by varying the number of target relations in the FB1.5M and Wiki-NYT dataset\footnote{The target relations are sampled three times randomly, and the results are averaged.}. Figure~\ref{fig:para2}  shows that when the number of target relations decreases, the performance of KG-BERT/CNN+DANN degrades quickly, implying the severe negative transfer. We observe that  RAN   outperforms KG-BERT/CNN+DANN when the number of target relations decreases. Note that, wRAN performs comparably to KG-BERT/CNN+DANN in standard-setting when the number of target relations is 60. This means that the weights mechanism will not wrongly filter out relations when there are no outlier relations.

\subsection{Case Study}
%We select  samples  from shared and outlier relations for detail analysis in \textbf{Case 1}, \textbf{Case 2} and \textbf{Case 3}. 

\textbf{Case 1: Relation-gate.}
We give some examples of how our relation-gate balance weights for relations and instances. In Table~\ref{tab:case123}, we display the $\alpha$ of different relations. 
For relation \emph{capital\_of}, there are lots of dissimilar relations, so relation weights are more important, which results in a small $\alpha$. 
For relation \emph{educated\_in}, the instance difference is more important so $\alpha$  is relatively large. 

\textbf{Case 2: Relation Correlation.}
We give some examples of how our approach assigns different weights for relations to mitigate the negative effect of outlier relations. In Table~\ref{tab:case123}, we display the sentences from shared relations and outlier relations. The relation \emph{capital\_of}  is an outlier relation whereas \emph{director} is a shared relation. We observe that our model can automatically find outlier relations and assign lower weights to them. 

\textbf{Case 3: Instance Transfer Ability.}
We give some examples of how our approach assigns different weights for instances to de-emphasize the nontransferable samples. In  Table~\ref{tab:case123}, we observe that: (1) Our model can automatically assign lower weights to instances in outlier relations. (2) Our model can assign different weights for instances in the same relation space to down-weight the negative effect of nontransferable instances.  
(3) Although our model can identify some nontransferable instances, it still assigns incorrect weights to some instances (The end row in Table~\ref{tab:case123}) which is semantically similar and transferable. We will address this by adopting additional mechanisms like transferable attention \cite{wang2019transferable}, which will be part of our future work.

\section{Conclusion and Future Work}

In this paper, we propose a novel model of \emph{Weighted  Relation Adversarial  Network} for low resource knowledge graph completion. %Our approach can make use of rich source relation information and identify unrelated relations/samples and down-weight their importance automatically to tackle the problem of negative transfer and encourage positive transfer.  
Extensive experiments demonstrate that our model achieves results that are comparable with that of state-of-the-art link prediction and relation extraction baselines. In the future, we plan to improve the results by jointly modeling link prediction and relation extraction. It will also be promising to apply our approach to other low resource NLP scenarios such as emerging event extraction, low resource question answering.

\section*{Acknowledgments}
We  want to express gratitude to the anonymous reviewers for their hard work. This work is funded by NSFC91846204/U19B2027 and national key research program SQ2018YFC000004 
%\clearpage
%\bigskip

\bibliography{WWW2020}
 
\bibliographystyle{ACM-Reference-Format}


\end{document}

\maketitle
\section{Theoretical Analysis}

After adding the importance weights to the source samples for the domain classifier $D_a$, the objective function of weighted domain adversarial nets $\mathbb{L}_w(D_a, F)$ is:
\begin{small}
\begin{equation}\begin{split}
\min \limits_{F_t} \max \limits_{D_a} \mathbb{L}(D,F_s,F_t) =& \mathbb{E}_{x \sim p_s(x)}[w^{total}logD(F_s(x))]+ \\&
\mathbb{E}_{x \sim p_t(x)}[1-logD(F_t(x))]
\label{eq: firstgan}
\end{split}
\end{equation}
\end{small}

where the $w^{total}$  is independent of $D_a$ and can be seen as a constant. Thus, given $F_s$ and $w^{total}$, for any $F_t$, the optimum $D_a$ of the weighted adversarial nets is obtained at:

\begin{equation}
D^{*}_a(x) = \frac{w^{total}p_s(x)}{w^{total}p_s(x)+p_t(x)}
\end{equation}

Note that since we normalized the importance weights $w^{total}$, the $w^{total}p_s(x)$ is still a probability density function:

\begin{equation}
\mathbb{E}_{x \sim p_s(x)}=  \int w^{total}p_s(x)dx = 1
\end{equation}

Given the optimum $D_a^*$, the minimax game of Equation \ref{eq: firstgan} can be reformulated as:
\begin{equation}
\begin{split}
 \mathbb{L}(F_t) =& \mathbb{E}_{x \sim p_s(x)}[w^{total}logD(F_s(x))]+ \\
&\mathbb{E}_{x \sim p_t(x)}[1-logD(F_t(x))] \\
= &  \int_x w^{total}p_s(x)log\frac{w^{total}p_s(x)}{w^{total}p_s(x)+p_t(x)}\\
&p_t(x)log\frac{(p_t(x))}{w^{total}p_s(x)+p_t(x)}\\
=& -log(4)+2 \dot JS(w_{total}p_s(x)||p_t(x))
\end{split}
\end{equation}

Hence, the weighted adversarial nets-based domain adaptation is essentially reducing the \emph{Jensen-Shannon divergence} between the weighted source density and the target density in the feature space, which obtains it’s optimum on $w^{total}p_s(x) = p_t(x)$.

\section{Computational Complexity}
The  additional computational complexity   of   our model  compared with  baselines are   the pretrain  of $F_s$  and $C$ to  get the category level weights  and pretrain  of $F_t$ and $D$ to get  instance level weights,  they have  computation complexity linear to the number of training samples.  Moreover, those weights can be shared or as initial values for different datasets (share  relations) to accelerate training.   In the test stage,  the  computation complexity  is the same as the baselines.  Note that, our model is useful for  the relation extraction task in domains which even don’t  have a pre-built  knowledge graph  for   distance supervision, e.g., relation extraction tasks of  financial  and   legal documents.

\section{Training Details}

To avoid the noisy signal at the early stage of training procedure, we use a similar schedule method as \cite{ganin2014unsupervised}
 for the tradeoff parameter to update $F_t$ by initializing it at 0 and gradually increasing to a pre-defined upper bound. The schedule is defined as: $\Phi = \frac{2 \cdot u}{1+ exp(-\alpha \cdot p)}-u$, where $p$ is the
the training progress linearly changing from 0 to 1, $\alpha$ = 1, and $u$ is the upper bound set to 0.1 in our experiments.

%We  apply virtual batch normalization (VBN) \cite{salimans2016improved} in the sentence encoder, in which each example $x$ is normalized based on the statistics collected on a reference batch of examples that are chosen once and fixed at the start of training, and on $x$ itself. The reference batch is normalized using only its statistics.

As  data imbalance will severely affect the model training and cause the model only sensitive to classes with a high proportion, we cut the propagation of loss from NA  to relieve the impact of NA followed by \cite{ye2017jointly}, and we set its loss as 0.

%More details.
 We implement our approach via Tensorflow \cite{abadi2016tensorflow} and adopt the Adam optimizer  \cite{kingma2014adam}. Table \ref{nyt} and Table \ref{ace} show our final setting for all hyper-parameters on the Wiki-NYT and ACE05 datasets respectively. We have uploaded  sampled training, validating and testing data  and publish the dataset online\footnote{https://github.com/zxlzr/WADARE/tree/master/data}. Baseline code can be found from Github\footnote{https://github.com/thunlp/OpenNRE}

\begin{table}[!htbp]
\centering

\begin{tabular}{c|c}

\hline
 \centering \textbf{Parameter}&  \textbf{Settings}\\
\hline
 \centering Kernel size $k$& 3 \\
 \centering Word embedding size&300  \\
\centering Position embedding size& 5 \\

\centering Number of Channels for PCNN& 230 \\
\centering Learning rate& 0.001 \\
\centering $\alpha$&0.9  \\
\centering $\beta$&0.6 \\

\centering $\gamma$&0.01  \\
\centering Dropout rate&0.5  \\
\centering Batch size&128  \\
\centering Epoch&30 \\

\hline
\end{tabular}\caption{Parameter settings on the Wiki-NYT dataset.}\label{nyt}

\end{table}
\begin{table}[!htbp]
\centering

\begin{tabular}{c|c}

\hline
 \centering \textbf{Parameter}&  \textbf{Settings}\\
\hline
 \centering Kernel size $k$& 3 \\
 \centering Word embedding size&300  \\
\centering Position embedding size& 5 \\
\centering Number of Channels for CNN& 150 \\
\centering Learning rate& 0.001 \\
\centering $\alpha$&0.5  \\

\centering $\lambda$&0 \\
\centering $\gamma$&0.001 \\
\centering Dropout rate&0.5  \\
\centering Batch size&128  \\
\centering Epoch&10 \\

\hline
\end{tabular}\caption{Parameter settings on the ACE05 dataset.}\label{ace}

\end{table}

%Moving Semantic Transfer Network

\begin{algorithm}[th]
\caption{Moving semantic transfer algorithm to compute $\mathbb{L}_{SM}$ in iteration $t$,  $K$ is the number of classes.}
\begin{algorithmic}[1]
\REQUIRE  Labeled set $S$, unlabeled set $T$, $N$ is the batch size, train classifier $f^* = C(F_s(x))$, global centroids for two domains: $\{C_s^k\}_{k=1}^K$ and $\{C_t^k\}_{k=1}^K$
\STATE   $S_t$ = RandomSample(S,N)
\STATE $T_t$ = RandomSample(T,N)
\STATE  $\hat T_t$ = Lalebling($f^*$,$T_t$)
\STATE  $\mathbb{L}_{SM}$ = 0
\FOR{$k = 1$ to K}
\STATE  $C^k_{s_{(t)}} \leftarrow \frac{1}{|S_t^k|}   \sum \limits_{(x_i,y_i) \in S_t^k} G(x_i)$ (From Scratch)
\STATE  $C^k_{t_{(t)}} \leftarrow \frac{1}{|\hat T_t^k|}   \sum \limits_{(x_i,y_i) \in \hat T_t^k} G(x_i)$ (From Scratch)
\STATE $C_s^k \leftarrow \zeta C_s^k + (1-\zeta)C^k_{s_{(t)}} $ (Moving Average)
\STATE $C_t^k \leftarrow \zeta C_t^k + (1-\zeta)C^k_{t_{(t)}}$ (Moving Average)
\STATE $\mathbb{L}_{SM} \leftarrow \mathbb{L}_{SM} + \Omega(C_s^kC_t^k)$
\ENDFOR
\RETURN  $\mathbb{L}_{SM}$
\end{algorithmic}\label{semantic}
\end{algorithm}

\section{Semantic Centroid Alignment}
We  apply the  moving semantic transfer algorithm followed by \cite{xie2018learning} to handle the false labels problem. Detailed algorithm is shown in Algorithm \ref{semantic} and source code available on Github\footnote{https://github.com/Mid-Push/Moving-Semantic-Transfer-Network}. 
Instead of aligning those newly obtained centroids in each iteration directly, we  align exponential moving average centroids. We maintain global centroids for each class. In each iteration, source centroids are updated by the labeled source samples while target centroids are updated by pseudo-labeled target samples. Then we can align those moving average centroids.

\bigskip

%Our model successfully obtains improvement on all three test domains of relations at ACE 2005. It uses a domain adversarial neural network to learn cross-domain features. It does not require handcrafted features for domain adaptation. It can be a useful tool for relation extraction since labeled data is always hard to acquire.

\bibliographystyle{acl_natbib}
\bibliography{aaai18}